\documentclass{article}

\usepackage[preprint, nonatbib]{main}

\usepackage[utf8]{inputenc} 
\usepackage[T1]{fontenc}    
\usepackage{hyperref}       
\usepackage{url}            
\usepackage{booktabs}       
\usepackage{amsfonts}       
\usepackage{nicefrac}       
\usepackage{microtype}     
\usepackage{xcolor}        
\usepackage{algorithm}
\usepackage{algorithmic}
\usepackage{bm}
\usepackage{booktabs} 
\usepackage{multirow}
\usepackage{multicol}
\usepackage{tabularx}
\usepackage{amsmath}
\usepackage{graphicx}
\usepackage{amssymb}
\usepackage{subfigure}
\usepackage{color}
\usepackage{graphicx}
\usepackage{wrapfig}

\title{A Bayesian Mixture Model of Temporal Point Processes with Determinantal Point Process Prior}

\author{
  Yiwei Dong \\
  Renmin University of China\\
  \And
  Shaoxin Ye \\
  Renmin University of China\\
  \And
  Yuwen Cao \\
  Renmin University of China\\
  \And
  Qiyu Han \\
  Renmin University of China\\
  \And
  Hongteng Xu \thanks{Corresponding authors.}\\
  Renmin University of China\\
  \And
  Hanfang Yang \footnotemark[1]\\
  Renmin University of China\\
}

\begin{document}

\maketitle

\begin{abstract}
Asynchronous event sequence clustering aims to group similar event sequences in an unsupervised manner. Mixture models of temporal point processes have been proposed to solve this problem, but they often suffer from overfitting, leading to excessive cluster generation with a lack of diversity. To overcome these limitations, we propose a Bayesian mixture model of \textbf{T}emporal \textbf{P}oint \textbf{P}rocesses with \textbf{D}eterminantal \textbf{P}oint \textbf{P}rocess Prior (\textbf{TP$^2$DP$^2$}) and accordingly an efficient posterior inference algorithm based on conditional Gibbs sampling. Our work provides a flexible learning framework for event sequence clustering, enabling automatic identification of the potential number of clusters and accurate grouping of sequences with similar features. It is applicable to a wide range of parametric temporal point processes, including neural network-based models. Experimental results on both synthetic and real-world data suggest that our framework could produce moderately fewer yet more diverse mixture components, and achieve outstanding results across multiple evaluation metrics. 
\end{abstract}

\section{Introduction}\label{sec:intro}

As a powerful tool of asynchronous event sequence modeling, the temporal point process (TPP) plays a crucial role in many application scenarios, such as modeling trading activities in financial markets~\cite{10.1145/3604237.3626893,10.1609/aaai.v37i6.25897}, analyzing information dissemination within social networks~\cite{Farajtabar2015COEVOLVEAJ,Li_Ke_2020,fang2023group}, and depicting the propagation of infectious diseases~\cite{9722957,pmlr-v28-yang13a}. 
Traditional TPPs such as the Poisson process~\cite{daley2003basic} and Hawkes process~\cite{hawkes1971spectra} offer straightforward and intuitive parametric models for building interpretable generative mechanisms of event sequences. 
Recently, TPPs with neural network architectures have been developed to capture the intricate relationship among events in real-world sequences, in which events may either excite or inhibit each other's occurrence and exhibit both short-term and long-term impacts. 
In particular, leveraging architectures like recurrent neural network (RNN)~\cite{mei2017neural,du2016recurrent}, attention mechanism~\cite{zhang2020self,zuo2020transformer,zhu2021deep,mei2022transformer}, and normalizing flow~\cite{shchur2019intensity,chen2020neural}, these neural TPPs have proven highly effective in capturing the complex temporal dynamics of event occurrences.

\begin{figure}
    \centering
    \includegraphics[width=0.69\linewidth]{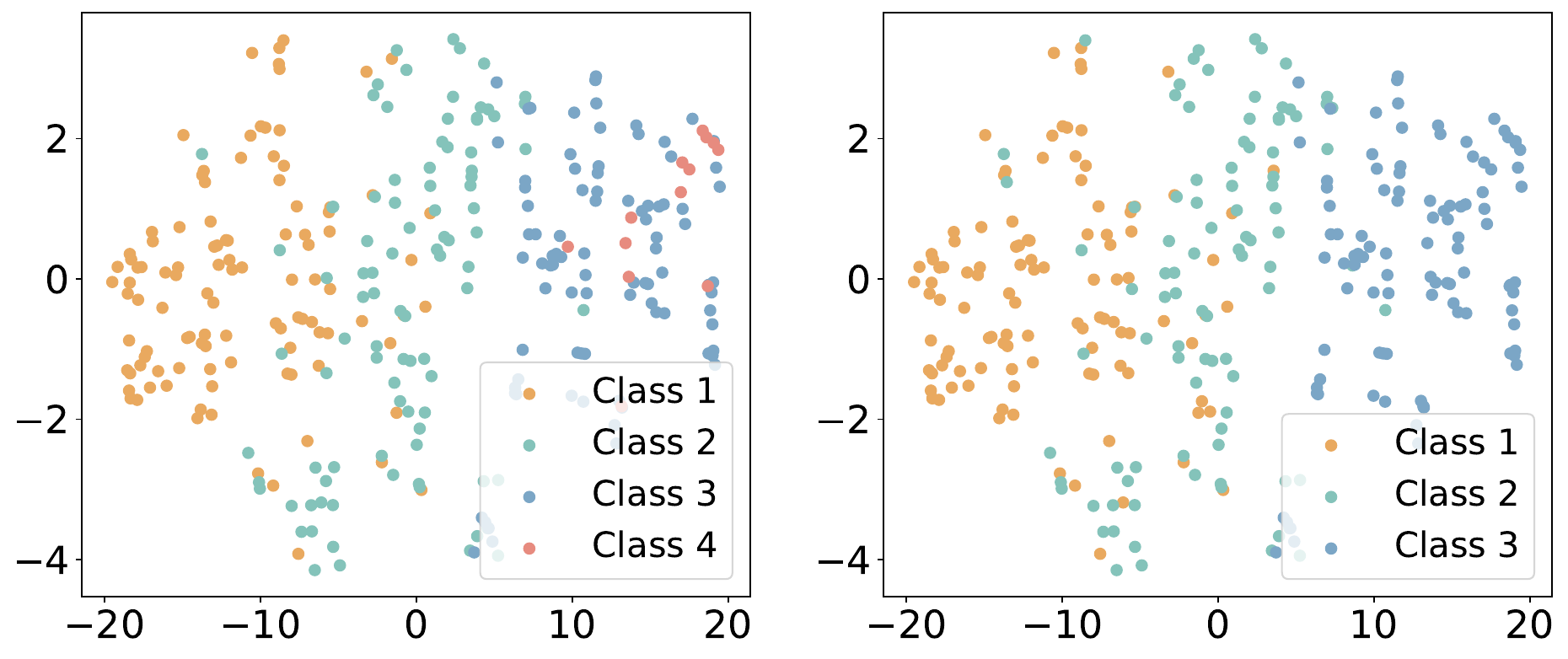}
    \caption{The t-SNE plots~\cite{van2008visualizing} of RMTPP's event sequence embeddings~\cite{du2016recurrent} for a synthetic dataset with three clusters. NTPP-MIX~\cite{zhang2022learning} (left) produces four clusters wrongly, while our TP$^2$DP$^{2}$ (right) leads to the clustering results matching well with the ground truth.}
    \label{fig:tsne}
\end{figure}

In practice, event sequences often demonstrate clustering characteristics, with certain sequences showcasing greater similarities when compared with others. 
For instance, event sequences of patient admissions may exhibit clustering patterns in response to specific medical treatments, and individuals sharing the same occupation often display similar shopping behaviors. 
Being able to accurately cluster event sequences would bring benefits to many fields, including medical health and personalized recommendations. 
In recent years, researchers have built mixture models of TPPs to tackle the event sequence clustering problem~\cite{NIPS2017_dd8eb9f2,9063463,zhang2022learning}. 
However, these models often suffer from overfitting during training, leading to excessive cluster generation with a lack of diversity. We illustrate this issue in the left panel of Figure~\ref{fig:tsne}. Moreover, these methods require either manually setting the number of clusters in advance~\cite{9063463} or initializing a large number of clusters and gradually removing excessive clusters through hard thresholding~\cite{NIPS2017_dd8eb9f2,zhang2022learning}.
In addition, without imposing proper prior knowledge, the clusters obtained by these models may have limited diversity and cause the identifiability issue.

\begin{figure}[t]
    \centering
    \includegraphics[width = 0.79\linewidth]{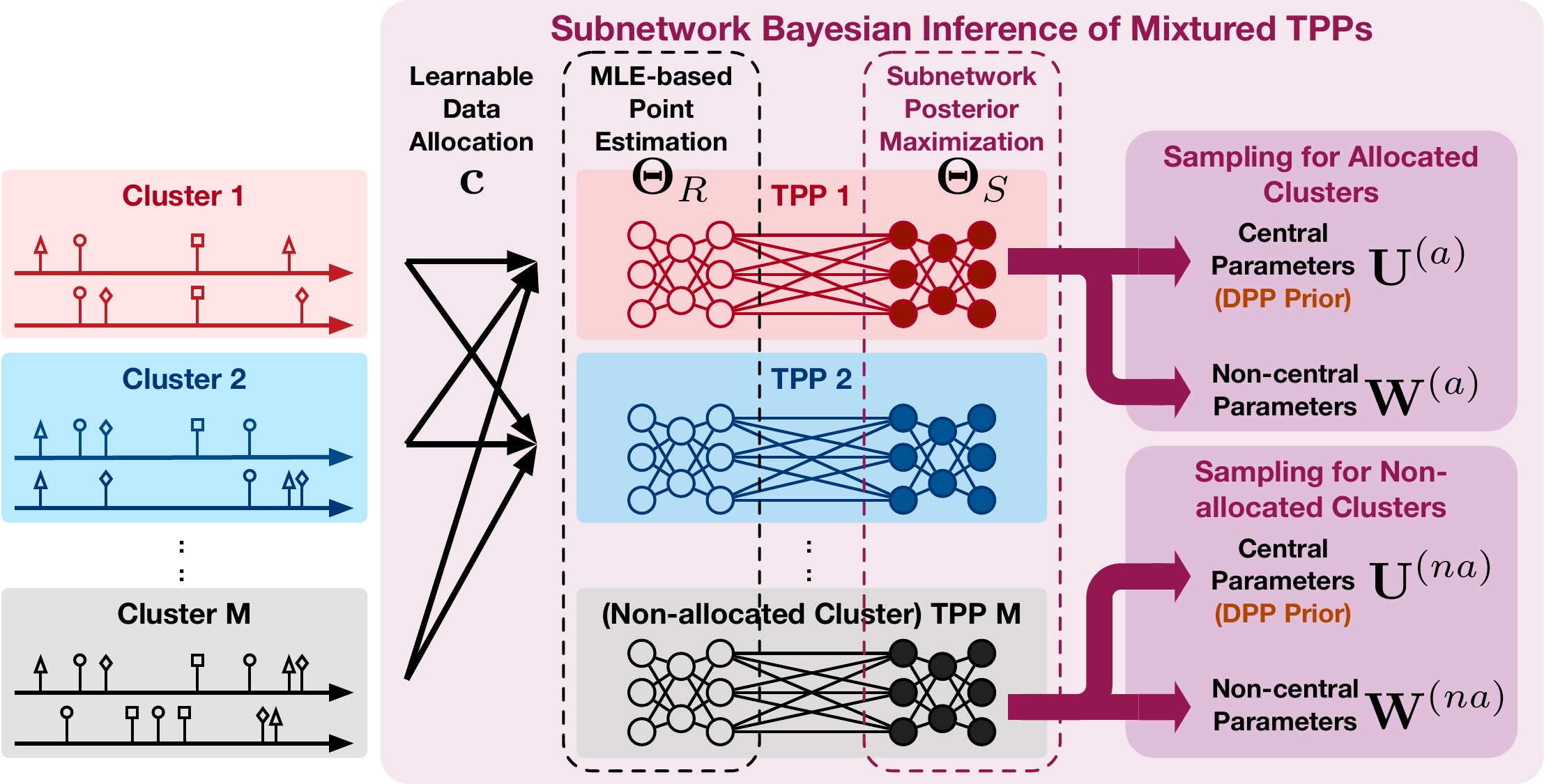}
    \caption{The scheme of our TP$^2$DP$^2$.  
    Some model parameters (i.e., $\bm{\Theta}_S$) apply Bayesian inference while the remaining ones (i.e., $\bm{\Theta}_R$) apply maximum likelihood estimation. 
    We split $\bm{\Theta}_S$ into central parameters that determine clustering structures and non-central parameters, respectively. 
    The DPP prior of the central parameters encourages diverse clusters without predefining the number of clusters.}
    \label{fig:dppppl}
\end{figure}

In this study, we propose a novel Bayesian mixture model of temporal point processes named TP$^2$DP$^2$ for event sequence clustering, imposing a determinantal point process prior to enhance the diversity of clusters and developing a universally applicable conditional Gibbs sampler-based algorithm for the model's posterior inference.
As illustrated in Figure~\ref{fig:dppppl}, TP$^2$DP$^2$ leverages the determinantal point process (DPP) as a repulsive prior for the parameters of cluster components, which contributes to generating TPPs with diverse parameters. 
To make TP$^2$DP$^2$ applicable for the TPPs with a large number of parameters, we apply Bayesian subnetwork inference~\cite{daxberger2021bayesian}, employing Bayesian inference to partially selected parameters while utilizing maximum likelihood estimation for the remaining parameters. 
For the selected parameters, we further categorize them into central and non-central parameters, in which the central parameters mainly determine the clustering structure and thus we apply the DPP prior. 
We design an efficient conditional Gibbs sampler-based posterior inference algorithm for the selected parameters, in which the stochastic gradient Langevin dynamics~\cite{welling2011bayesian} is introduced into the updating process to facilitate convergence. 

To our knowledge, TP$^2$DP$^{2}$ is the first work that explores event sequence clustering based on a TPP mixture model with determinantal point process prior. 
It automatically identifies the number of clusters, whose clustering results are more reliable than existing variational inference methods~\cite{NIPS2017_dd8eb9f2,zhang2022learning}, as shown in Figure~\ref{fig:tsne}. 
We apply TP$^2$DP$^{2}$ to cluster synthetic and real-world event sequences and compare its performance with state-of-the-art methods. 
Extensive experiments demonstrate that TP$^2$DP$^{2}$ applies to classic and neural network-based backbones and achieves superior cluster detection capabilities.

\section{Preliminaries \& Related Work}\label{sec2}
\paragraph{Temporal Point Processes}
TPP is a kind of stochastic process that characterizes the random occurrence of events in multiple dimensions, whose realizations can be represented as event sequences, i.e., $\{(t_i, d_i)\}_{i=1}^{I}$, where $t_i \in [0,T]$ are time stamps and $d_i\in \mathcal{D}=\{1,...,D\}$ are different dimensions (a.k.a. event types). 
Typically, we characterize a TPP by conditional intensity functions:
\begin{eqnarray}
    \lambda^*(t) = \sideset{}{_{d=1}^D}\sum\lambda_d^*(t),~\text{and}~\lambda_d^*(t)\mathrm{d}t={\mathbb{E}[\mathrm{d}N_d(t) \mid \mathcal{H}_t]}.
\end{eqnarray}
Here, $\lambda_d^*(t)$ is the conditional intensity function of the type-$d$ event at time $t$, $N_d(t)$ denotes the number of the occurred type-$d$ events prior to time $t$, and $\mathcal{H}_t$ denotes the historical events happening before time $t$.
Essentially, $\lambda_d^*(t)$ represents the expected instantaneous rate of event occurrence at time $t$ given the past event information. 

The simplest TPP is Poisson process~\cite{daley2003basic}, whose events happen independently at a constant rate, i.e., $\lambda_d^*(t) = \lambda_d$, $\forall d\in\mathcal{D}$. 
Hawkes process~\cite{hawkes1971spectra} introduces triggering patterns among different event types, whose conditional intensity function is $\lambda_d^{*}(t) = \mu_d + \sum_{t_i<t} \alpha_{dd_i} e^{-\beta(t-t_i)}$, where $\mu_d$ represents the base intensity of the type-$d$ event, $\alpha_{dd_i}$ encodes the infectivity of the type-$d_i$ event on the type-$d$ event, and $\beta$ controls the decay rate of the infectivity over time. 
Neural TPPs set $\{\lambda_d^{*}(t)\}_{d=1}^{D} = f_{\theta}(t,\mathcal{H}_t)$, where $f_{\theta}$ is a neural network parameterized by $\theta$, which can be recurrent neural networks~\cite{mei2017neural,du2016recurrent} or Transformer architectures~\cite{zhang2020self,zuo2020transformer,mei2022transformer}. 
Given an event sequence $\bm{s} = \{(t_i, d_i)\}_{i=1}^{I}$, the likelihood function of a TPP can be derived based on its conditional intensity functions: 
\begin{eqnarray}\label{eq:like}
\begin{aligned}
    \mathcal{L}(\bm{s}) = \sideset{}{_{i=1}^{I}}\prod \lambda_{d_i}^*(t_i)  \exp\Bigl(-\int_0^T \lambda^*(\tau) d\tau\Bigr).
\end{aligned}
\end{eqnarray}
By maximizing the likelihood in Eq.~\eqref{eq:like}, we can learn the TPP model to fit the observed sequence.

\paragraph{Mixture Models of TPPs} 
Given multiple event sequences belonging to different clusters, i.e., $\{\bm{s}_n\}_{n=1}^{N}$, we often leverage a mixture model of TPPs to describe their generative mechanism, leading to a hierarchical sampling process:
\begin{eqnarray}\label{eq:mixture0}
\begin{aligned}
    &\text{1) Determine cluster: }m\sim \text{Categorical}(\bm{\pi}), \text{2) Sample sequence: }\bm{s} \sim \text{TPP}({\bm{\theta}_m}),
\end{aligned}
\end{eqnarray}
where $\bm{\pi}=[\pi_1,...,\pi_M]\in\Delta^{M-1}$ indicates the distribution of clusters defined on the ($M-1$)-Simplex, $\text{TPP}({\bm{\theta}_m})$ is the TPP model corresponding to the $m$-th cluster, whose parameters are denoted as $\bm{\theta}_m$. 
In~\cite{NIPS2017_dd8eb9f2}, the Dirichlet mixture of Hawkes processes is proposed and is the first model-based clustering method for event sequences, yet the mixture components of this model are restricted to Hawkes processes. 
The work in~\cite{zhang2022learning} extends the mixture model for neural TPPs. 
A reinforcement learning-based clustering algorithm has also been proposed~\cite{9063463} for neural TPPs, but it only applies to unidimensional event sequences. 
In addition, these methods require setting a large cluster number in advance, along with the imposition of a hard threshold to gradually remove excessive clusters throughout the training process. 
As mentioned above, they often suffer from overfitting and lead to excessive and non-diverse cluster generation. 

\paragraph{Determinantal Point Processes}\label{dpppara}

DPP~\cite{kulesza2012determinantal} is a stochastic point process characterized by the unique property that its sample sets exhibit determinantal correlation. 
The structure of DPP is captured through a kernel function~\cite{lavancier2015determinantal,bianchini2020determinantal}, which determines the similarity or dissimilarity between points in the sample space. 
Denote the kernel function by $\kappa:\mathcal{X}\times \mathcal{X}\mapsto\mathbb{R}$, where $\mathcal{X}$ represents a sample space.
The probability density function for samples  $x_1,...,x_M\in \mathcal{X}$ in one realization of DPP is: 
\begin{equation}\label{dppeq}
\begin{aligned}
    p(x_1,...,x_M)\propto {\mathrm{det}}\{\bm{K}(x_1,...,x_M)\},
\end{aligned}
\end{equation}
where $\bm{K}(x_1,...,x_M)=[\kappa(x_i,x_{j})]$ is a $M \times M$ Gram matrix corresponding to the samples. 

Given arbitrary two samples $x_i$ and $x_j$, we have $p(x_i, x_j)=\kappa(x_i,x_i)\kappa(x_j,x_j)-\kappa(x_i,x_j)^2=p(x_i)p(x_j)-\kappa(x_i,x_j)^2 \leq p(x_i)p(x_j)$.
Therefore, DPP manifests the repulsion between $x_i$ and $x_j$ --- a large value of $\kappa(x_i,x_j)$ implies that $x_i$ and $x_j$ are unlikely to appear together. 
This correlation property makes DPP suitable for modeling repulsion effects and sampling diverse data subsets. 
In particular, using DPP as the prior in Gaussian mixture models~\cite{xu2016bayesian,xie2019bayesian,bianchini2020determinantal,beraha2022mcmc} often helps enhance the diversity of clustering results. 

To our knowledge, none of the existing work considers the utilization of the DPP prior in the mixture model of TPPs.

\section{Proposed TP$^2$DP$^2$ Model}\label{sec3}

The mixture model in Eq.~\eqref{eq:mixture0} reveals that each event sequence $\bm{s}$ obeys a mixture density, i.e., $\sum_{m=1}^{M} \pi_m \mathcal{L}(\bm{s} \mid \bm{\theta}_m)$, where $M$ is a random variable denoting the number of clusters, $\bm{\pi}=[\pi_1,...,\pi_M]\in\Delta^{M-1}$ specifies the probability of each cluster component (a TPP), and $\mathcal{L}(\bm{s} \mid \bm{\theta}_m)$ is the likelihood of the $m$-th TPP parametrized by $\bm\theta_m$. 
Given $N$ event sequences $\bm{S}=\{\bm{s}_n\}_{n=1}^{N}$, we denote cluster allocation variables of each sequence $\bm{c}=[c_1,...,c_N]\in\{1,...,M\}^N$, where each set $\{\bm{s}_n \mid c_n=m\}$ contains the sequences assigned to the $m$-th cluster. 
Accordingly, we derive the joint distribution of all variables, i.e., $p(M, \bm{\Theta}, \bm{\pi}, \bm{c}, \bm{S})$, as 
\begin{eqnarray}\label{ppdf5}
\begin{aligned}
p(M)p(\bm\Theta  \mid  M)  p(\bm\pi  \mid  M)\underbrace{p(\bm{c}  \mid  \bm{\pi})p(\bm{S}  \mid  \bm{\Theta}, \bm{c})}_{\prod_{n=1}^N \pi_{c_n}\mathcal{L}(\bm{s}_n  \mid  \bm\theta_{c_n})},
\end{aligned}
\end{eqnarray}
where $\bm\Theta=\{\bm{\theta}_m\}_{m=1}^{M} \in \mathbb{R}^P$. 
$p(M)$, $p(\bm\Theta  \mid  M)$ and $p(\bm\pi  \mid  M)$ are prior distributions of $M$, $\bm\Theta$ and $\bm\pi$, respectively. 
By Bayes Theorem, the posterior distribution $p(M, \bm\Theta, \bm\pi, \bm{c}  \mid  \bm{S} )$ is proportional to Eq.~\eqref{ppdf5}.

The exact sampling from $p(M, \bm\Theta, \bm\pi, \bm{c}  \mid  \bm{S} )$ is often intractable because the parameters of the TPPs in practice (especially those neural TPPs~\cite{mei2017neural,du2016recurrent,zhang2020self,zuo2020transformer,mei2022transformer}) are too many to perform full Bayesian posterior calculation. 
To overcome this issue, we conduct posterior inference only on a subset of model parameters (i.e., the ``subnetwork'' of the whole model). 
In particular, we approximate the full posterior of the TPPs' parameters $\bm{\Theta}$ as
\begin{eqnarray}\label{approxeq}
\begin{aligned}
p(\bm\Theta  \mid  \bm{S}) &\approx p\left(\bm{\Theta}_S  \mid  \bm{S}\right) \delta(\bm{\Theta}_R-\widehat{\bm{\Theta}}_R) = p(\bm{U}  \mid  \bm{S})p(\bm{W} \mid \bm{S})\delta(\bm{\Theta}_R-\widehat{\bm{\Theta}}_R), 
\end{aligned}
\end{eqnarray}
where we split the model parameters $\bm{\Theta}$ into two parts, i.e., $\bm{\Theta}_S$ and $\bm{\Theta}_R$, respectively. 
$\bm\Theta_S=\{\bm{\theta}_{S,m}\}_{m=1}^{M}$ corresponds to the subnetworks of the TPPs in the mixture model, which we set to be last few layers in nerual TPPs, while $\bm{\Theta}_R=\{\bm{\theta}_{R,m}\}_{m=1}^{M}$ denotes the remaining parameters.
In Eq.~\eqref{approxeq}, $p(\bm\Theta  \mid  \bm{S})$ is decomposed into the posterior of the subnetworks $p(\bm\Theta_S \mid \bm{S})$ and a Dirac delta function on the remaining parameters $\bm{\Theta}_R$, in which $\bm{\Theta}_R$ is estimated by their point estimation $\widehat{\bm\Theta}_R=\{\widehat{\bm{\theta}}_{R,m}\}_{m=1}^{M}$, e.g., the maximum likelihood estimation achieved by stochastic gradient descent. 
Theories and experiments have proven that a partial Bayesian network can still ensure model expressiveness and predictive performance and even outperforms their full Bayesian counterparts~\cite{daxberger2021bayesian,sharma2023bayesian,izmailov2020subspace}.

Unlike existing work, in Eq.~\eqref{approxeq}, we further decompose the parameters in the subnetworks into two parts, i.e., $\bm{\Theta}_S=\{\bm{U},\bm{W}\}$, where $\bm{U}=\{\bm{\mu}_m\}_{m=1}^{M}$ and $\bm{W}=\{\bm{w}_m\}_{m=1}^{M}$, respectively.
For the $m$-th TPP in the mixture model, $\bm{\mu}_m$ corresponds to the ``central'' parameters of their conditional intensity functions, which significantly impacts the overall dynamics of event occurrence. 
In Hawkes processes~\cite{hawkes1971spectra}, the base intensity signifies the traits of the background process. 
In neural TPPs~\cite{mei2017neural,du2016recurrent,zhang2020self,zuo2020transformer,mei2022transformer}, the bias term of their last linear layer functions in a similar way.
Therefore, we denote each $\bm{\mu}_m$ as the base intensity in Hawkes process and the bias term of the last linear layer in neural TPP, respectively.\footnote{
Assuming the exchangeability in $p(\bm{\mu}_1, \cdots, \bm{\mu}_{\Tilde{M}})$ for any fixed number $\Tilde{M}$, then $\bm U$ is a finite point process which specifies both the random cluster number $M$ and the value of central parameters. 
Additionally, we have $\bm U \in \Omega =\cup_{m=0}^\infty\Omega_m$, where $\Omega_m$ denotes the space of all finite subsets of cardinality $m$. 
The related measure theoretical guarantee of $\bm U$ can be found in our Appendix.} 
Accordingly, the other ``non-central'' parameters in each subnetwork are denoted as $\bm{w}_m$, which are contingent upon specific architectures of different models.

Imposing the conditional independence on the central and non-central parameters, i.e.,  $p(\bm{\Theta}_S | M)=p(\bm{U} | M)p(\bm{W} | M)$, we have
\begin{eqnarray}\label{ppdf}
\begin{aligned}
p(M, \bm\Theta, \bm\pi, \bm{c}  |  \bm{S} ) \propto p(M)p(\bm{U}  |  M)p(\bm{W} | M)  p(\bm{\pi}  |  M) \sideset{}{_{n=1}^N}\prod \pi_{c_n}\mathcal{L}(\bm{s}_n  |  \bm{\theta}_{S,c_n},\widehat{\bm\theta}_{R,c_n}),
\end{aligned}
\end{eqnarray}
where $\widehat{\bm\theta}_{R,c_n}$ denotes the point estimates of the remaining parameters in the $c_n$-th TPP. 
We set the priors of key parameters (i.e., $M$, $\bm{U}$, $\bm{W}$, and $\bm{\pi}$) as follows: 

\textbf{Prior of $M$:} We let the prior of $M$ be the uninformative prior~\cite{beraha2022mcmc} supported on all positive integers, instead of making it fixed. 
    Thus, we may leave out $M$ in the following context for brevity.
    
\textbf{Prior of $\bm{U}$:} DPP prior is introduced to the central parameter $\bm{U}$ to mitigate the overfitting problem and diversify the cluster result. 
    We leverage the spectral density approach to approximate the DPP density. 
    For a DPP shown in Eq.~\eqref{dppeq}, its kernel function has a spectral representation  $\kappa(\bm{\mu}_i,\bm{\mu}_j)=\sum_{i=1}^\infty \lambda_i\phi_i(\bm{\mu}_i)\overline{\phi_i(\bm{\mu}_j)}$, in which each eigenfunction could be approximated via the Fourier expansion and eigenvalues are specified by the spectral distribution, as defined in~\cite{lavancier2015determinantal}. 
    In this way, the DPP density approximation is
    \begin{eqnarray*}
       p(\bm U  \mid  M) \approx  \exp \left( |\mathcal{R}| -D_{\text {app}}\right) \operatorname{det}\{\tilde{\bm{K}}\}(\bm \mu_1, \cdots, \bm \mu_M),
    \end{eqnarray*}
    where $\tilde{{\kappa}}(\bm{\mu}_i, \bm{\mu}_j)=\sum_{\bm z \in \mathbb{Z}^q} \tilde{\varphi}(\bm z) \mathrm{e}^{2 \pi \mathrm{i} \bm z \cdot(\bm{\mu}_i-\bm{\mu}_j)}$, $D_{\text {app}}=\sum_{\bm z \in \mathbb{Z}^q} \log (1+\tilde{\varphi}(\bm z))$, $\{\bm \mu_1, \cdots, \bm \mu_M\} \subset \mathcal{R}$, $|\mathcal{R}|$ is the volume of the range of the parameter space, $\tilde{\varphi}(\bm z)=\varphi(\bm z) /(1-\varphi(\bm z))$, $\mathbb{Z}^q$ is $q$-dimensional integer lattice, and $\varphi$ is the spectral distribution. 
    
\textbf{Prior of $\bm{W}$:} 
    We have $p(\bm{W} \mid M)=\prod_{m=1}^{M}p(\bm{w}_m)$, where $p(\bm{w}_m)$ is the prior of the $m$-th TPP's non-central parameter. 
    In this study, we use exponential distribution priors when the $\bm w$'s are the triggering parameters in Hawkes processes and Gaussian priors for the parameters in neural networks. 
    
\textbf{Prior of $\bm{\pi}$:} Instead of directly sampling  $\{\pi_m\}_{m=1}^M$ from its posterior distribution, we apply the ancillary variable method~\cite{beraha2022mcmc,10.1214/22-AOS2201} to make the posterior calculation tractable for the mixture weights $\{\pi_m\}_{m=1}^M$. 
    Consider $\bm{r}=[r_1,...,r_M]$, which consists of i.i.d. positive continuous random variables following the Gamma distribution $\Gamma(1, 1)$, each $r_m$ is independent of $M$ and $\bm{r}$ is independent of $\{\bm{U},\bm{W}\}$. 
    Defining $t=\sum_{m=1}^Mr_m$ and $\bm{\pi}=[r_1/t,...,r_M/t]$, we establish a one-to-one correspondence between $\bm{\pi}$ and $(\bm r,t)$. 
    By introducing an extra random variable $v \sim \Gamma(N, 1)$, we define the ancillary variable $u = v/t$, with 
    \begin{equation}
    p(u) = \frac{u^{N-1}}{\Gamma(N)}\int_0^\infty t^N e^{-ut} p(t)\mathrm{d}t.
    \end{equation}
    Introducing $u$ makes the posterior computation of $\bm{\pi}$ factorizable and gets rid of the sum-to-one constraint imposed on $\{\pi_m\}_{m=1}^M$, significantly simplifying the subsequent MCMC simulation process. 

In summary, the joint posterior density function becomes
\begin{eqnarray}
\begin{aligned}
    p(M, \bm\Theta, \bm{c}, \bm{r}, u  \mid  \bm{S} ) \propto p(\bm{U} )\sideset{}{_{m=1}^{M}}\prod p(\bm{w}_m)p(r_m)  \sideset{}{_{n=1}^N} \prod \pi_{c_n}\mathcal{L}(\bm{s}_n  \mid  \bm{\theta}_{S,c_n},\widehat{\bm\theta}_{R,c_n})\frac{p(u \mid t)}{t^N},
\end{aligned}
\label{eqnonsplit}
\end{eqnarray}
where $p(\bm{U})=p(M)p(\bm{U} \mid M)$ is the DPP prior which determines the central parameters and the corresponding cluster numbers jointly. 

\section{Posterior Inference Algorithm} 

Because TP$^2$DP$^2$ have a large number of parameters in general, we first use the stochastic gradient descent algorithm to pretrain the model and then use conditional Gibbs samplers to infer the parameters of the selected subnetworks. 
Since the number of clusters changes dynamically as these algorithms proceed, it is helpful to further partition model parameters into parameters of allocated clusters and those of non-allocated clusters when applying posterior sampling. 
In particular, we partition $\bm{U}$ into two sets according to cluster allocations $\bm c$: one comprising cluster centers currently used for data allocation, denoted as $\bm{U}^{(a)} = \{\bm\mu_{c_1}, \ldots, \bm\mu_{c_n}\}$, and the other containing cluster centers not involved in the allocation, denoted as $\bm{U}^{(na)} = \bm{U} \setminus \bm{U}^{(a)}$. 
Note that the product measure $\mathrm{d}\bm{\mu} \times \mathrm{d}\bm{\mu}$ in $\Omega \times \Omega$ lifted by the map $(\bm{x}, \bm{y}) \mapsto \bm{x} \cup \bm{y}$ results in the measure $\mathrm{d}\bm{\mu}$, so the prior density of $(\bm{U}^{(a)}, \bm{U}^{(na)})$ is equivalent to $p(\bm{U}^{(a)},\bm{U}^{(na)})=p(\bm{U}^{(a)}\cup\bm{U}^{(na)})$, which follows the DPP density. 
$\bm W$ and $\bm r$ are partitioned in the same way.

As $(\bm U,\bm \pi,\bm W,\bm c)$ and 
$(\bm{U}^{(a)},\bm r^{(a)},\bm W^{(a)},\bm{U}^{(na)},\bm r^{(na)},$ $\bm W^{(na)},\bm c)$ are in a one-to-one correspondence, we can refer to Eq.~\eqref{eqnonsplit} and obtain the posterior of $(M, \bm{U}^{(a)},\bm r^{(a)},\bm W^{(a)},\bm c, \bm{U}^{(na)},\bm r^{(na)}, \bm W^{(na)}, u)$ as 
\begin{eqnarray}\label{eq:postnew}
\begin{aligned}
&p(M, \bm{U}^{(a)},\bm r^{(a)},\bm W^{(a)},\bm c,\bm{U}^{(na)},\bm r^{(na)},\bm W^{(na)}, u  |  \bm{S})\\
\propto & p(\bm{U}^{(a)}\cup\bm{U}^{(na)}) \Bigl[\sideset{}{_{m=1}^k}\prod p(\bm w^{(a)}_m) p(r^{(a)}_m)(r^{(a)}_m)^{n_m} \sideset{}{_{i: c_i = m}}\prod \mathcal{L}(\bm s_i  \mid  \bm{\mu}^{(a)}_m, \bm w^{(a)}_m,\widehat{\bm\theta}_{R,m}) \Bigr] \\ & \times \Bigl[\sideset{}{_{m=1}^l}\prod p(\bm w^{(na)}_m) p(r^{(na)}_m)\Bigr] p(u  \mid  t) \frac{1}{t^N},  
\end{aligned}
\end{eqnarray}
where $n_m$ is the number of sequences allocated to the $m$-th component, $k$ denotes the cardinality of allocated clusters, and $l$ denotes that of non-allocated ones. $r_m^{(a)}$ and $r_m^{(na)}$ denote the allocated and non-allocated unnormalized weight, respectively.

Our posterior inference algorithm follows the principle of conditional Gibbs sampler~\cite{papaspiliopoulos2008retrospective}. 
We split all parameters into three groups: an allocated block $(\bm{U}^{(a)},\bm r^{(a)},\bm W^{(a)})$, a non-allocated block $(\bm{U}^{(na)},\bm r^{(na)},\bm W^{(na)})$, and remaining parameters $\{\bm c, u\}$, and update them in an alternating scheme. 

\paragraph{Sampling non-allocated variables:}
We begin with sampling $\bm U^{(na)}$ from its conditional density:
\begin{eqnarray}\label{e:bbb}
 \begin{aligned}
     & \quad \quad p(\bm U^{(na)}  \mid \bm U^{(a)}, \bm r^{(a)}, \bm W^{(a)}, \bm c, u, \bm{S}) \\ 
     &=\iint p(\bm U^{(na)}, \bm r^{(na)}, \bm W^{(na)}  \mid  \cdots) \, \mathrm{d}\bm r^{(na)} \, \mathrm{d}\bm W^{(na)}\\
     &\propto \iint p(\bm{U}^{(a)}\cup\bm{U}^{(na)})
 \Bigl[\sideset{}{_{m=1}^l}\prod p(\bm w^{(na)}_m) p(r^{(na)}_m)\Bigr] \\ &\times p(u  |  t) \frac{1}{t^N} \, \mathrm{d}\bm r^{(na)} \, \mathrm{d}\bm W^{(na)} 
 = p(\bm{U}^{(a)}\cup\bm{U}^{(na)}) \psi(u)^l,
\end{aligned}
\end{eqnarray}
where ``$\cdots$'' denotes variables excluding the target variable to be sampled, together with all the sample sequences $\bm{S}$, and henceforth.  
$p(\bm U^{(a)} \cup \bm U^{(na)} )$ is the DPP density. 
The second term $\psi(u)^l  = \int[\prod_{m=1}^l \exp(-u r^{(na)}_m) p(r^{(na)}_m)]  \mathrm{d}\bm r^{(na)}$, due to the fact that
\begin{eqnarray}
\begin{aligned}
&p(u \mid t) = \frac{u^{N-1}}{(N-1)!} e^{-ut} t^{N} = \frac{t^{N}u^{N-1}}{(N-1)!} \Bigl[\sideset{}{_{m=1}^k}\prod e^{-u r^{(a)}_m}\Bigr]\Bigl[ \sideset{}{_{m=1}^l}\prod e^{-u r^{(na)}_m}\Bigr].
\end{aligned}
\end{eqnarray}
Applying Birth-and-death Metropolis-Hastings algorithm~\cite{geyer1994simulation}, we sample $\bm U^{(na)}$ and determine the final number of clusters accordingly. 

We then sample $\bm r^{(na)}$ and $\bm W^{(na)}$ using the classical Metropolis-Hastings algorithm. 
The cardinality of non-allocated variables (i.e., $l$) is determined by the size of $\bm U^{(na)}$, so we have
\begin{equation}\label{sna}
\begin{aligned}
   &p(\bm r^{(na)}  \mid  \cdots) \propto \sideset{}{_{m=1}^l}\prod p(r^{(na)}_m) e^{-u r^{(na)}_m},  \\ &p(\bm W^{(na)}  \mid  \cdots)\propto \sideset{}{_{m=1}^l}\prod p(\bm w^{(na)}_m).
\end{aligned}
\end{equation}

\paragraph{Sampling allocated variables:}
The allocated central parameter $\bm U^{(a)}$ is sampled from
\begin{eqnarray}\label{mua}
\begin{aligned}
         &p(\bm{U}^{(a)}  \mid  \cdots) \propto p(\bm U^{(a)} \cup \bm U^{(na)} )   \sideset{}{_{m=1}^k}\prod \sideset{}{_{i: c_i = m}}\prod \mathcal{L}(\bm s_i  \mid  (\bm{\mu}^{(a)}_m, \bm w^{(a)}_m, \widehat{\bm\theta}_{R,m}),
\end{aligned}
\end{eqnarray}
where the $p(\bm U^{(a)} \cup \bm U^{(na)} )$ is again governed by the DPP. 
Subsequently, we sample $\bm r^{(a)}$ from its full conditional using the Metropolis-Hastings algorithm:
\begin{eqnarray}\label{sa}
    p(\bm r^{(a)} \mid \cdots) \propto \sideset{}{_{m=1}^k}\prod p(r^{(a)}_m) (r^{(a)}_m)^{n_m}  \exp(-u r^{(a)}_m). 
\end{eqnarray}
The $\bm W^{(a)}$'s full conditional is:
\begin{eqnarray}\label{wa}
\begin{aligned}
    &p(\bm W^{(a)} \mid \cdots) \propto \sideset{}{_{m=1}^k}\prod p(\bm w^{(a)}_m)  \sideset{}{_{i: c_i = m}}\prod \mathcal{L}(\bm s_i \mid (\bm{\mu}^{(a)}_m,\bm w^{(a)}_m, \widehat{\bm\theta}_{R,m})).
\end{aligned}
\end{eqnarray}
As $\bm W^{(a)}$ represents all the allocated parameters of the point process model to be inferred, excluding $\bm \mu$, it may still exhibit high dimensionality. 
To align with our framework and boost convergence, we leverage the stochastic gradient Langevin dynamics~\cite{welling2011bayesian} when sampling $\bm W^{(a)}$. 
The proposed update for each $\bm w_m^{(a)}$ is provided by:
\begin{eqnarray}
 \begin{aligned}
    &\Delta \bm w^{(a)}_{m} = \eta_j + \frac{\epsilon_j}{2} \Bigl( \nabla \log p(\bm w^{(a)}_m)   +\frac{n_m}{n_*} \sideset{}{_{c_i = m}}\sum\nabla \log \mathcal{L}(\bm s_i  \mid  \bm{\mu}^{(a)}_m,\bm w^{(a)}_m, \widehat{\bm\theta}_{R,m}) \Bigr),
 \end{aligned}
\end{eqnarray}
where $j$ is the counting number of iterations, $\eta_j \sim N\left(0, \epsilon_j\right)$, $\epsilon_j$ is the step size at the $j$-th iteration which is set to decay towards zero, and $n_*$ in the above equation is the number of selected sequences from each cluster to perform stochastic approximation. 
$\nabla \log \mathcal{L}(\bm s_i  \mid  \bm{\mu}^{(a)}_m,\bm w^{(a)}_m, \widehat{\bm\theta}_{R,m}) $ is calculated through the automatic differentiation~\cite{baydin2018automatic}.

\paragraph{Sampling $\bm{c}$ and $u$:}
Each cluster label $c_i$ is sampled from 
\begin{eqnarray}\label{ci}
\begin{aligned}
    p(c_i = m\mid\cdots)\propto
    \begin{cases}
       & r^{(a)}_m \mathcal{L}(\bm s_i  \mid  \bm{\mu}^{(a)}_m,\bm w^{(a)}_m, \widehat{\bm\theta}_{R,m})   \quad \text{for~}m=1,...,k,\\
        &  r^{(na)}_m \mathcal{L}(\bm s_i  \mid  \bm{\mu}^{(na)}_m,\bm w^{(na)}_m, \widehat{\bm\theta}_{R,m})   \quad\text{for~}m=k+1,...,k+l.
    \end{cases}
\end{aligned}
\end{eqnarray}
Note that after this step, there is a positive probability that $c_i > k$ for certain indices $i$, indicating that some initially non-allocated components become allocated, and vice versa---some initially allocated components become non-allocated. Consequently, a relabeling of $(\bm{U}^{(a)}, \bm r^{(a)}, \bm W^{(a)}, \bm{U}^{(na)}, \bm r^{(na)}, \bm W^{(na)})$ and $\bm c$ is performed, ensuring that $\bm c$ takes values within the set $\{1, \ldots, k\}^N$. Thus, $k$ may either increase or decrease or remain unchanged after the relabeling step. 
Finally, we sample $u$ from a gamma distribution with a shape parameter of $N$ and an inverse scale parameter of $t$. 
We summarize the complete procedure of the posterior inference algorithm of TP$^2$DP$^2$ in Appendix \ref{suppc}.

\section{Experiments}

To comprehensively evaluate the effectiveness of our TP$^2$DP$^2$ model and its inference algorithm, we test our method on both synthetic and real-world datasets and compare it with state-of-the-art event sequence clustering methods.
For each method, we evaluate its clustering performance by clustering purity~\cite{NIPS2017_dd8eb9f2} and adjusted rand index (ARI)~\cite{vinh2009information} and its data fitness by the expected log-likelihood per event (ELL)~\cite{xue2024easytpp}.

In addition, we report the expected posterior value of the number of clusters ($M$) in real-world dataset, which reveals the inferred number of components given data. 
The code of TP$^2$DP$^2$ is available at \url{https://anonymous.4open.science/r/TP2DP2/}.

\subsection{Experiments on Mixtures of Hawkes Processes}

We first investigate the clustering capability of  TP$^2$DP$^2$ and demonstrate the rationality of DPP priors on the synthetic datasets generated by mixture models of Hawkes processes, in which each Hawkes process generates $100$ event sequences with 3 event types. 
All Hawkes processes apply the same triggering function, and their base intensities are set to $\bm{\mu}_m=(0.5+\delta_m)\bm{1}_3$, where $\bm{1}_3$ is a three-dimensional column vector, with each element equal to 1. $\delta_m=\delta\cdot(m-1)$ for $m\in\{1,2,3,\cdots, K_{GT}\}$, where ${K_{GT}}$ denotes the true number of clusters. In other words, these Hawkes processes exhibit distinct temporal patterns because of their different base intensities. The experiments are carried out for \({K_{GT}} = 4, 5\) for multiple datasets, each dataset having different $\delta$ values, $ \delta \in \{0.1, 0.15, 0.2, 0.25, 0.3, \cdots, 1\}$.

\begin{figure}[t]
    \centering
    \begin{minipage}[b]{0.48\linewidth}
    \centering
\includegraphics[width=0.79\linewidth]{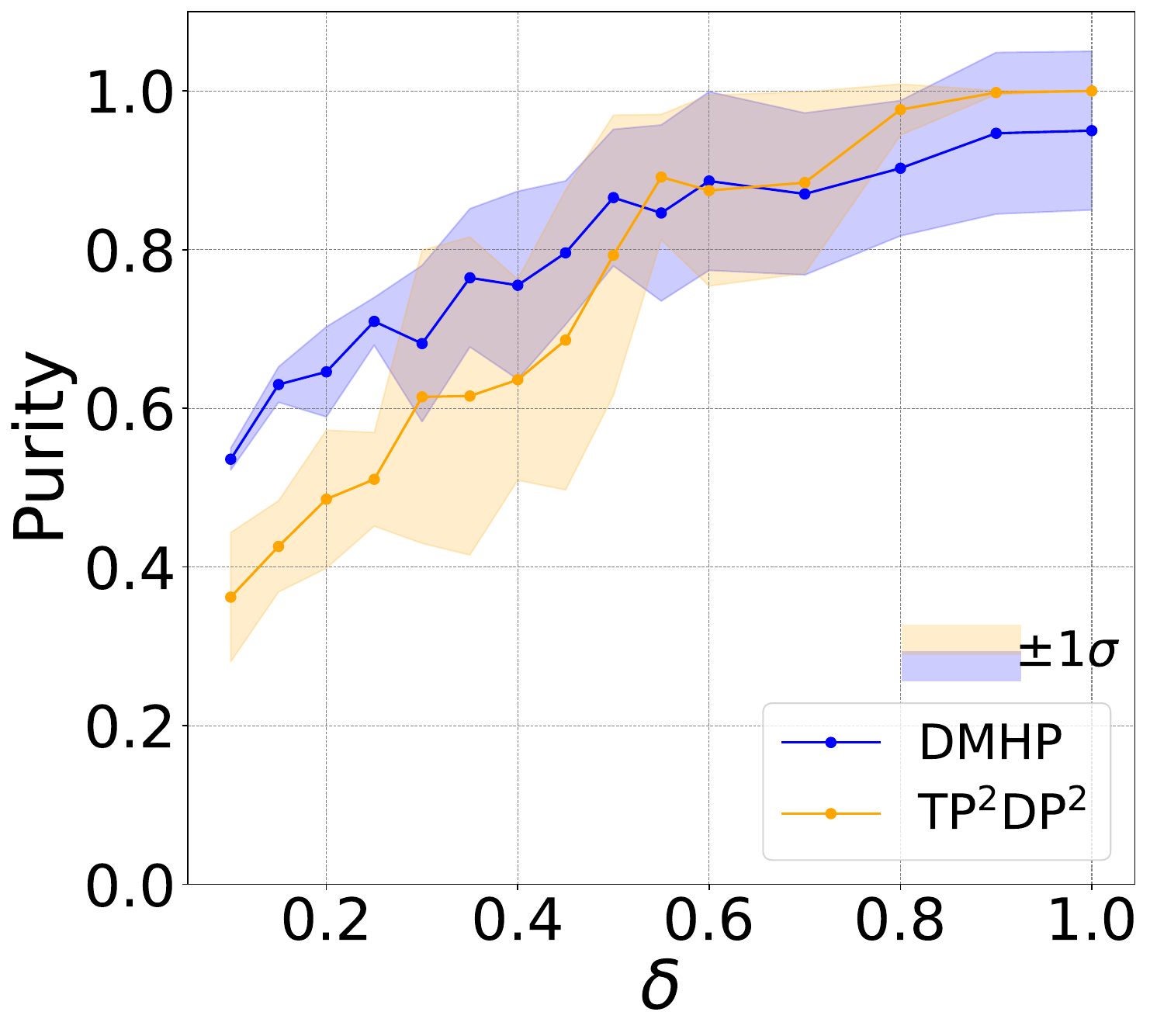}
        \label{fig:subfig3}
    \end{minipage}
    \hspace{-0.8cm}
    \begin{minipage}[b]{0.48\linewidth}
      \centering
\includegraphics[width=0.79\linewidth]{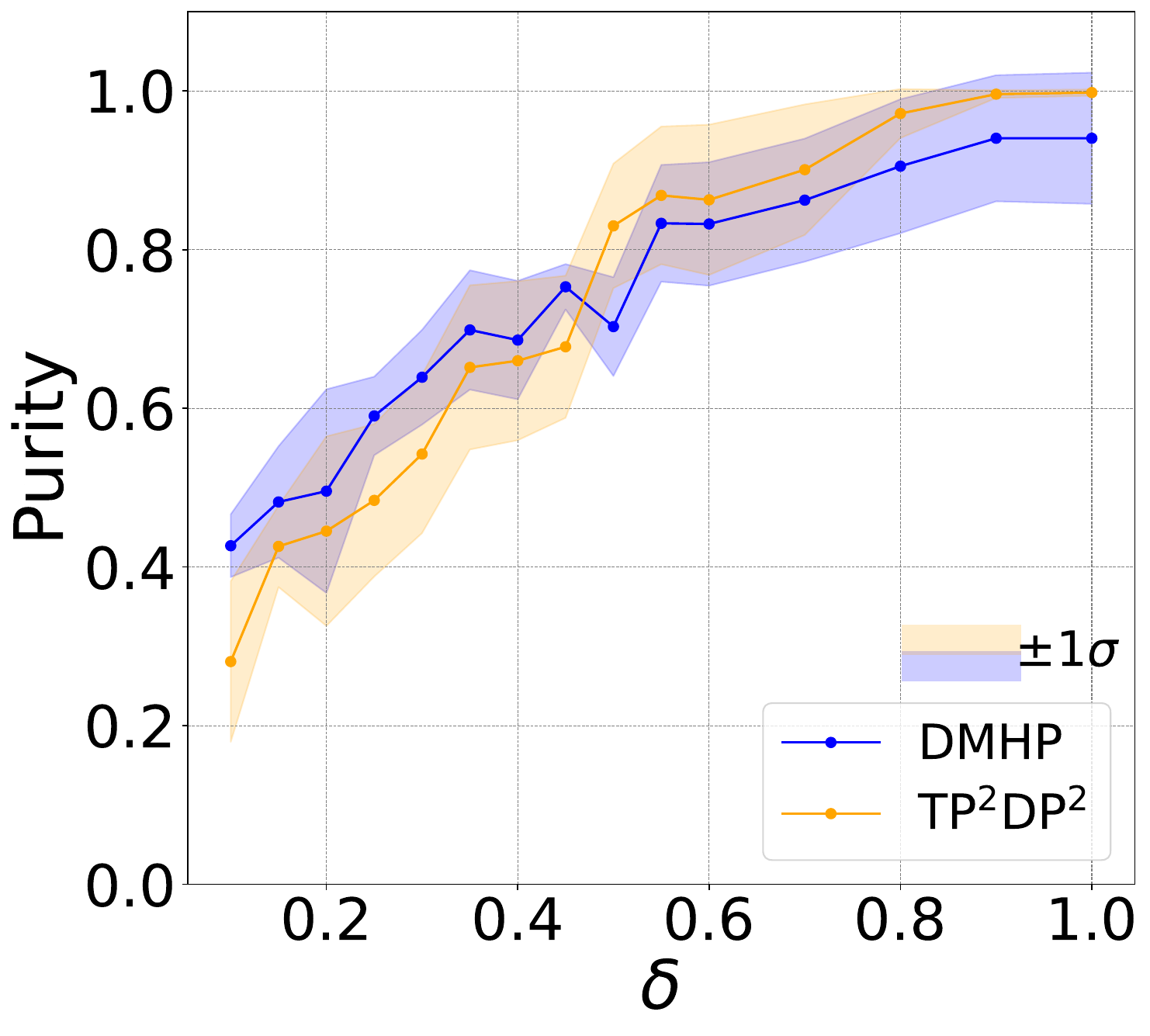}
        \label{fig:subfig4}
    \end{minipage}
    \caption{The means and standard deviations of clustering purity obtained by DMHP and TP$^2$DP$^2$ with different $\delta$. 
    The left panel is the result when the ground truth cluster number $K_{GT} = 4$, and the right is the result of $K_{GT} = 5$.}
    \label{fig:sy1figures}
\end{figure}

We compare our TP$^2$DP$^2$ with the Dirichlet mixture model of Hawkes processes (\textbf{DMHP}) learned by the variational EM algorithm~\cite{NIPS2017_dd8eb9f2}.  
For a fair comparison, we use the same Hawkes process backbone as in DMHP, ensuring identical parametrization, and we consider all model parameters in the Bayesian inference phase. 
For each method, we initialize the number of clusters randomly in the range $[K_{GT}-1, K_{GT}+1]$. The averaged results in five trials are presented in Figure~\ref{fig:sy1figures}.  

When the disparity in the true base intensity among different point processes is minimal, the inherent distinctions within event sequences are not apparent, as shown in Figure~\ref{fig:tsne_syn1_true}. 
In this case, TP$^2$DP$^2$ tends to categorize these event sequences into fewer groups than the ground truth, resulting in a relatively modest purity when $\delta$ is small. 
As $\delta$ increases, TP$^2$DP$^2$ exhibits increasingly robust clustering capabilities, consistently outperforming DMHP when $\delta > 0.55$.

\begin{figure}[t]
    \centering
    \includegraphics[width=0.50\linewidth]{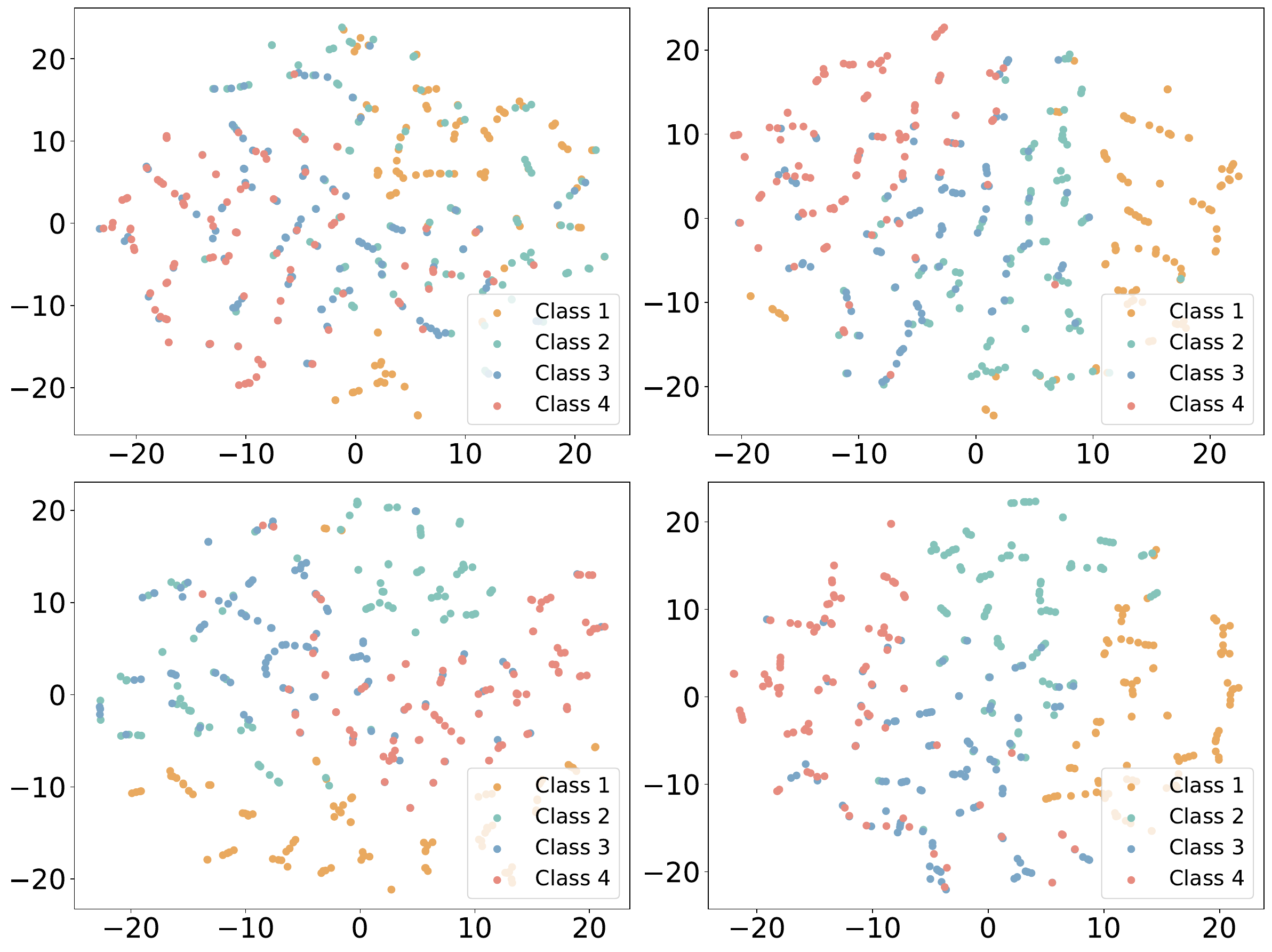}
    \caption{The t-SNE plot of the ground truth distribution for the synthetic mixture of Hawkes processes datasets with $\delta$ values of 0.2 (upper left), 0.4 (upper right), 0.6 (lower left), and 0.8 (lower right).}
    \label{fig:tsne_syn1_true}
\end{figure}

\begin{figure}[t]
    \centering
    \includegraphics[width = 0.55\linewidth]{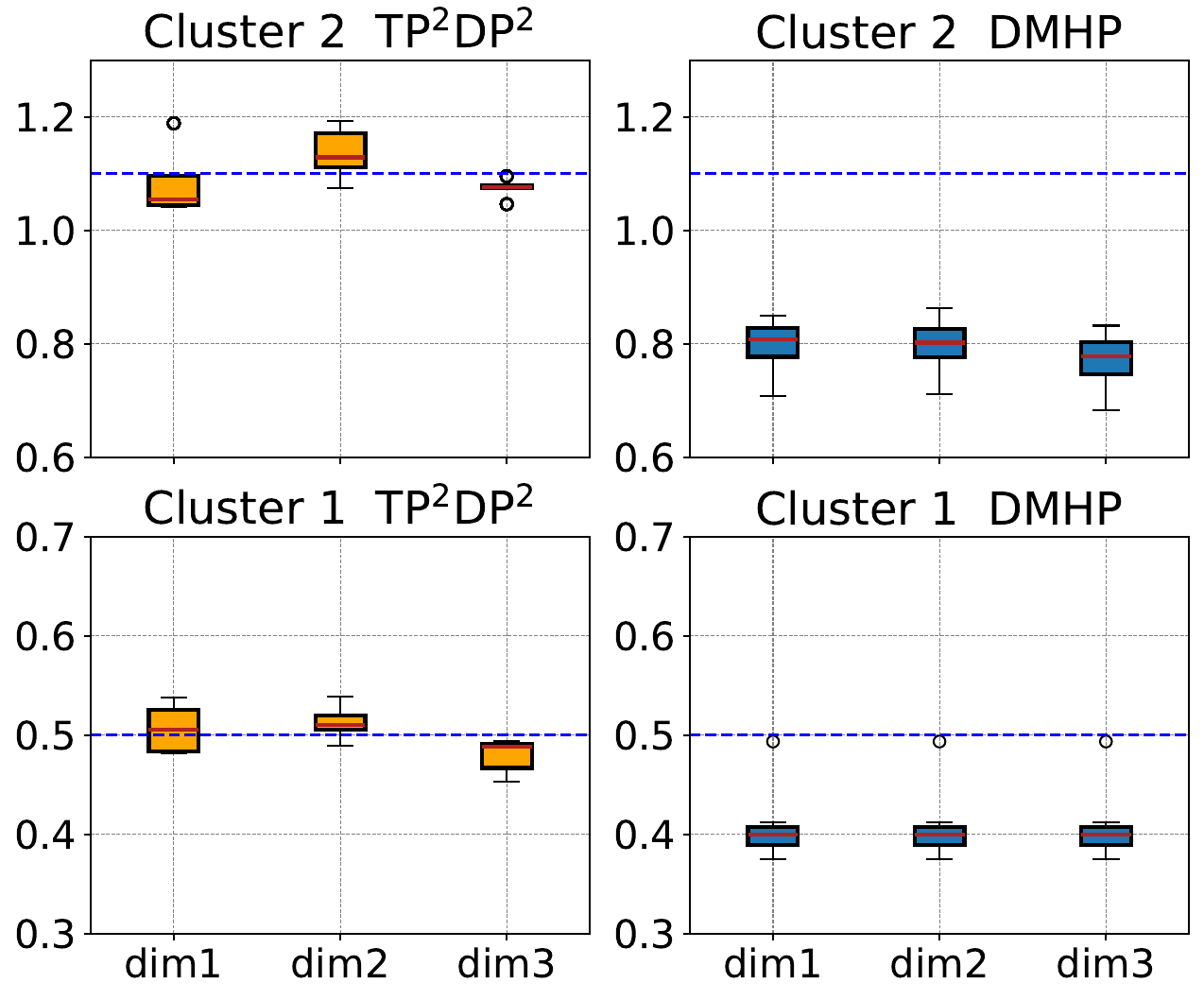}
    \caption{The base intensity $\bm\mu$ of first two clusters learned by two methods across 5 random trials.  
    The dotted line represents the ground truth $\bm\mu$ in two clusters.}
    \label{fig:box}
\end{figure}

In addition, we examine the posterior distribution of base intensity parameters when both algorithms converge. 
At $\delta = 0.6$, box plots in Figure~\ref{fig:box} depict posterior estimations of base intensities for the first two clusters (the ground truth base intensities are 0.5 and 1.1). 
It is noteworthy that DMHP consistently underestimates the true values in all trials due to multiple times of approximations in its learning algorithm, and DMHP shows marginal disparity between clusters. 
In contrast, TP$^2$DP$^2$ better captures the true base intensity values, meantime exhibiting greater dispersion between clusters compared with DMHP. 
Similar patterns are also observed in other datasets.

\begin{table*}[t]
  \centering
  \caption{Experimental results on synthetic mixture of hybrid point processes datasets.}
  \resizebox{\textwidth}{!}{
    \begin{tabular}{c|c|cc|cc|cc}
    \toprule
    \multirow{2}{*}{Backbone} & 
    \multirow{2}{*}{Method} & 
    \multicolumn{2}{c|}{$K_{GT}=3$} &
    \multicolumn{2}{c|}{$K_{GT}=4$} &
    \multicolumn{2}{c}{$K_{GT}=5$} \\
    &      
    & 
    Purity & 
    ARI & 
    Purity & 
    ARI &
    Purity & 
    ARI \\
    \midrule
    \multirow{2}[0]{*}{Hawkes} & 
    Dirichlet Mixture   & 
    $\text{0.678}_{\pm \text{0.134}}$  &  
    $\text{0.622}_{\pm \text{0.097}}$ &
    $\text{0.620}_{\pm \text{0.120}}$      &  
    $\text{0.564}_{\pm\text{0.126}}$ &
    $\text{0.574}_{\pm\text{0.045}}$      &  
    $\textbf{0.545}_{\pm \textbf{0.046}}$ \\
    & 
    TP$^2$DP$^2$   &   
    $\textbf{0.884}_{\pm\textbf{0.009}}$   &  
    $\textbf{0.745}_{\pm\textbf{0.052}}$ &
    $\textbf{0.739}_{\pm\textbf{0.004}}$    & 
    $\textbf{0.626}_{\pm\textbf{0.008}}$ &
    $\textbf{0.603}_{\pm\textbf{0.008}}$     & 
    $\text{0.538}_{\pm\text{0.013}}$ \\
    \midrule
    \multirow{2}[0]{*}{RMTPP} & 
    Dirichlet Mixture   &  
    $\textbf{0.983}_{\pm\textbf{0.112}}$     &  
    $\textbf{0.972}_{\pm\textbf{0.124}}$ &
    $\text{0.751}_{\pm\text{0.131}}$     &  
    $\textbf{0.712}_{\pm\textbf{0.213}}$ &
    $\text{0.708}_{\pm\text{0.030}}$     &  
    $\text{0.633}_{\pm\text{0.027}}$ \\
    & 
    TP$^2$DP$^2$   & 
    $\text{0.974}_{\pm\text{0.073}}$  &  
    $\text{0.971}_{\pm\text{0.109}}$  &
    $\textbf{0.753}_{\pm\textbf{0.003}}$     & 
    $\text{0.708}_{\pm\text{0.014}}$ &
    $\textbf{0.732}_{\pm\textbf{0.024}}$     &  
    $\textbf{0.674}_{\pm\textbf{0.017}}$ \\
    \midrule
    \multirow{2}[0]{*}{THP} & 
    Dirichlet Mixture   &  
    $\text{0.941}_{\pm \text{0.093}}$     & 
    $\text{0.870}_{\pm \text{0.201}}$   &
    $\text{0.746}_{\pm \text{0.007}}$     & 
    $\textbf{0.666}_{\pm \textbf{0.038}}$  &
    $\text{0.610}_{\pm \text{0.007}}$     &  
    $\text{0.559}_{\pm \text{0.043}}$ \\
    & 
    TP$^2$DP$^2$   &   
    $\textbf{0.980}_{\pm\textbf{0.035}}$     &   
    $\textbf{0.897}_{\pm\textbf{0.110}}$ &
    $\textbf{0.749}_{\pm\textbf{0.002}}$      &  
    $0.652_{\pm\textbf{0.007}}$ &
    $\textbf{0.650}_{\pm\text{0.007}}$      & 
    $\textbf{0.600}_{\pm\textbf{0.020}}$ \\
    \bottomrule
    \end{tabular}
    }
  \label{tbsyn2}
\end{table*}
\subsection{Experiments on Mixtures of Hybrid TPPs}

Our model is compatible with various TPP backbones, which can detect clusters and fit event sequence data originating from a mixture of hybrid TPPs. 
To verify our claim, we generate a set of event sequences based on five different TPPs, including 1) Homogeneous Poisson process, 2) Inhomogeneous Poisson process, 3) Self-correcting process, 4) Hawkes process, and 5) Neural Hawkes process~\cite{mei2017neural}. 
Based on the sequences, we construct three datasets with the number of mixture components ranging from three to five. 
For each dataset, we learn a mixture model of TPPs and set the backbone of the TPPs to be 1) the classic Hawkes process~\cite{hawkes1971spectra}, 2) the recurrent marked temporal point process (RMTPP)~\cite{du2016recurrent}, and 3) the Transformer Hawkes process (THP)~\cite{zuo2020transformer}, respectively.
The learning methods include the variational EM of Dirichlet mixture model~\cite{zhang2022learning} and our TP$^2$DP$^2$.
The results in Table~\ref{tbsyn2} show that our method achieves competitive performance. 
Especially when the backbone is Hawkes process, applying our method leads to notable improvements, which means that our method is more robust to the model misspecification issue.
In addition, learning RMTPP and THP by our method results in the best performance when $K_{GT}=5$, showcasing TP$^2$DP$^2$'s adaptability to complex event sequences.

We further investigate the effect of incorporating DPP prior to different parameters in the models, and the results are shown in Table~\ref{tbabl2}. In this experiment, we aim to verify adding DPP priors to central parameters of TPPs would lead to superior performance. For Hawkes Process, the base intensity reflects the average level of event occurrence rate, and is considered the most crucial for analyzing the feature of corresponding event sequences~\cite{hawkes1971spectra}. For example, in the field of seismology, the base intensity  specifically relates to background seismicity that needs special attention. Thus, for the event sequence clustering task, we also add DPP priors to the base intensity of the Hawkes process components, and find this yields the best clustering performance. For both RMTPP and THP, adding DPP prior to the output linear layer bias terms achieves the best purity and ARI scores, which are also consistently higher than the Dirichlet mixture frameworks or models without DPP priors. According to the architecture of these two neural point processes, the bias term of the last output layer has a direct impact on the estimated intensity function. 
This experimental result shows the effectiveness of applying DPP to the parameters that play a decisive role in intensity.

\begin{table}[h]
  \centering
  \caption{Experimental results on the synthetic mixture of hybrid point processes dataset ($K_{GT}=4$) when adding DPP prior to different parameters. None denotes we do not impose DPP prior.}
  \begin{tabular}{cccc}
    \toprule
    \multirow{1}{*}{Model} & \multirow{1}{*}{Layer} & Purity & ARI \\
    \midrule
            \multirow{3}{*}{Hawkes} & None & $\text{0.702}$ & $\text{0.648}$ \\  
                           &  Diagonal Elements of Infectivity Matrix& $\text{0.655}$ & $\text{0.572}$ \\
                           & Base Intensity & $\textbf{0.739}$ & $\textbf{0.626}$ \\
                     \midrule         
    \multirow{3}{*}{RMTPP} & None & $\text{0.750}$ & $\text{0.679}$ \\  
                           & Time Embedding Layer & $\text{0.747}$ & $\text{0.664}$ \\
                           & Output Layer & $\textbf{0.753}$ & $\textbf{0.708}$ \\
    \midrule
    \multirow{6}{*}{THP}   & None & $\text{0.722}$ & $\text{0.605}$ \\  
                           & Post-attention Feedforward Layer 1 & $\text{0.740}$ & $\text{0.630}$ \\
                           & Post-attention Feedforward Layer 2 & $\text{0.738}$ & $\text{0.610}$ \\
                           & Post-attention Feedforward Layer 3 & $\text{0.745}$ & $\text{0.647}$ \\
                           & Post-attention Feedforward Layer 4 & $\text{0.748}$ & $\text{0.647}$ \\
                           & Output Layer & $\textbf{0.749}$ & $\textbf{0.652}$ \\
    \bottomrule
  \end{tabular}
  \label{tbabl2}
\end{table}

\subsection{Experiments on Real-World Datasets}

To examine the performance of our method on real-world data, we use the following two benchmark datasets: 
1) Amazon~\cite{10.5555/3600270.3602780}. 
This dataset comprises time-stamped user product review events spanning from January 2008 to October 2018, with each event capturing the timestamp and the category of the reviewed product. Data is pre-processed according to the procedure in~\cite{xue2024easytpp}. The final dataset consists of 5,200 most active users with 16 distinct event types, and the average sequence length is 70. 
2) BookOrder \footnote{\label{note2}\url{https://ant-research.github.io/EasyTemporalPointProcess/user_guide/dataset.html}}. 
This book order dataset comprises 200 sequences, with two event types in each sequence. 
The sequence length varies from hundreds to thousands.

\begin{table}[t]
  \centering
  \caption{Experimental results on real-world datasets.}
  \small{
  \begin{tabular}{cccccc}
    \toprule
    \multirow{2}[0]{*}{Backbone} & \multirow{2}[0]{*}{Method} & \multicolumn{2}{c}{Amazon} & \multicolumn{2}{c}{BookOrder} \\
    \cmidrule(lr){3-4} \cmidrule(lr){5-6} 
          &       & \multicolumn{1}{c}{ELL} & \multicolumn{1}{c}{$M$} & \multicolumn{1}{c}{ELL} & \multicolumn{1}{c}{$M$} \\
    \midrule 
    \multirow{2}[0]{*}{Hawkes} & Dirichlet Mixture   &  -2.355 &  5.0   &  \textbf{4.832} &  3.6  \\
          & TP$^2$DP$^2$   &  $\textbf{-2.352}$  &   5.0    &    4.810  & 3.0 \\
    \midrule
    \multirow{2}[0]{*}{RMTPP} & Dirichlet Mixture   &   -2.251  &  3.8   &    5.613  &  2.6 \\
          & TP$^2$DP$^2$   & $\textbf{-2.052}$  &  3.0   &   \textbf{5.624}  &   2.2 \\
    \midrule
    \multirow{2}[0]{*}{THP} & Dirichlet Mixture   & 1.629    &   3.0  &    $5.814$   &  $2.6$   \\
          & TP$^2$DP$^2$   &  \textbf{1.631}  &  2.8  &  $\textbf{5.981}$  & $2.4$ \\
    \bottomrule
    \end{tabular}
    }
  \label{tbr1}
\end{table}

To ensure fairness, hyperparameters of the Dirichlet mixture models are tuned first and we intentionally make each backbone TPP model in TP$^2$DP$^2$ smaller or equivalent in scale compared to those of the Dirichlet mixture framework, which means that all hyperparameters related to the backbone structure, such as hidden size, number of layers, and number of heads within the TP$^2$DP$^2$ framework are set to be less than or equal to their corresponding counterparts in the Dirichlet mixture framework. 
In this case, if TP$^2$DP$^2$ model achieves a higher log-likelihood with fewer cluster numbers, it indicates that our method is better at capturing the characteristics of the data and provides a better fit.
Table~\ref{tbr1} summarizes the average results for different models in five trials.

In the experiment on the real-world dataset,  the Dirichlet Mixture framework performs generally worse than TP$^2$DP$^2$ in both dataset, but the number of posterior cluster numbers inferred by the Dirichlet Mixture framework is generally larger. This reflects that TP$^2$DP$^2$ moderately reduces the number of clusters to obtain more dispersed components without sacrificing much fitting capability.

\section{Conclusion}

In this paper, we propose the Bayesian mixture model TP$^2$DP$^2$ for event sequence clustering. 
It is shown that TP$^2$DP$^2$ could flexibly integrate various parametric TPPs including the neural network-based TPPs as components, achieve satisfying event sequence clustering results and produce more separated clusters. 
In the future, we plan to study the impact of alternative repulsive priors on event sequence clustering, and develop event sequence clustering methods in high-dimensional and spatio-temporal scenarios. It is also meaningful to explore the effect of DPP on mixture of intensity-free TPP models.

\bibliographystyle{plain}

{
\small
\bibliography{main}
}

\newpage
\appendix

The appendix consists of supplementary material for both theoretical analysis and experiments. 
In Appendix \ref{suppa}, we present the derivation of the log-likelihood function of the temporal point processes leveraging the intensity function.
In Appendix \ref{suppb}, we elaborate on the measure theoretical details of our method and the construction of DPP priors. In Appendix \ref{suppc}, we provide the complete procedure of the TP$^2$DP$^2$ algorithm. We provide the supplementary information for numerical experiments in Appendix \ref{suppd} and Appendix \ref{suppe}, including dataset details, implementation details, and additional results. 

\section{Derivation of the posterior of TP$^2$DP$^2$}\label{suppa}
\subsection{Derivation of the Likelihood Function for Temporal Point Processes} \label{app:llder}

Temporal Point Processes (TPPs) are stochastic processes that characterize the random occurrences of events over multiple dimensions, where realizations can be represented as event sequences $\{(t_i, d_i)\}_{i=1}^{I}$. Here, $t_i \in [0, T]$ denotes timestamps and $d_i \in \mathcal{D} = \{1, \ldots, D\}$ represents different event types. To describe the characteristics of a TPP, we typically employ the conditional intensity function:
$$
\lambda^*(t) = \sum_{d=1}^{D} \lambda_d^*(t), \quad \text{where} \  \lambda_d^*(t)\mathrm{d}t = \mathbb{E}[\mathrm{d}N_d(t)  \mid  \mathcal{H}_t].
$$
Here, $\lambda_d^*(t)$ denotes the conditional intensity function for event type $d$ at time $t$, $N_d(t)$ indicates the count of type-$d$ events up to time $t$, and $\mathcal{H}_t$ represents the history of events prior to time $t$. Essentially, $\lambda_d^*(t)$ captures the expected instantaneous rate of occurrence of type-$d$ events at time $t$, given the information of the previous events.

To derive the likelihood function of temporal point processes, we first need to establish the survival function $S(t)$, which gives the probability that no events occur up to time $t$. Assuming that the event time $T$ is a random variable with a probability density function $f_T(t)$, the cumulative distribution function $F_T(t)$ represents the probability that an event occurs before time $t$, i.e.
$
S(t) = 1 - F_T(t).
$
Given the definition of the conditional intensity function, the probability that an event occurs in a small interval $\mathrm{d}t$ can be approximated by $\lambda^*(t) \mathrm{d}t$. Hence, the probability that no events occur in the small interval $\mathrm{d}t$ following time $t$ is $1 - \lambda^*(t) \mathrm{d}t$. This implies that the probability of no events occurring up to time $t + \mathrm{d}t$ can be expressed as
$
S(t + \mathrm{d}t) = S(t) (1 - \lambda^*(t) \mathrm{d}t).
$

Taking the limit as $\mathrm{d}t \to 0$, we obtain the differential equation:

$$
\frac{dS(t)}{dt} = -\lambda^*(t) S(t).
$$

To this differential equation, we separate variables and integrate both sides:

$$
\frac{dS(t)}{S(t)} = -\lambda^*(t) \, dt, \ 
\int \frac{dS(t)}{S(t)} = -\int \lambda^*(t) \, dt,
$$
and then we have
$$ 
\log S(t) = -\int_{0}^{t} \lambda^*(u) \, du + C,
$$

where $C$ is the integration constant. Using the initial condition $S(0) = 1$, we find $C = 0$. Thus, the survival function is
$S(t) = \exp\left(-\int_{0}^{t} \lambda^*(u) \, du\right)
$, which expresses the probability that no events occur up to time $t$. Consequently, the probability that no events occur in the interval $[t_i, t_{i+1})$ is
 $\exp\left(-\int_{t_i}^{t_{i+1}} \lambda^*(u) \, du\right).
$

Next, we derive the likelihood function for the entire event sequence. The probability density of observing each event $(t_i, d_i)$ is given by the conditional intensity function $\lambda_{d_i}^*(t_i)$. Thus, the likelihood of observing the event sequence $\{(t_i, d_i)\}_{i=1}^{I}$ is the product of these probabilities $
\prod_{i=1}^{I} \lambda_{d_i}^*(t_i).
$

To account for the probability that no events occur between observed events, we incorporate the survival probabilities for the intervals $[t_i, t_{i+1})$, i.e.
$
\prod_{i=0}^{I} \exp\left(-\int_{t_i}^{t_{i+1}} \lambda^*(u) \, du\right),
$

where $t_0 = 0$ and $t_{I+1} = T$. Combining these, the likelihood function for the event sequence is:

$$
\mathcal{L}(\bm{s}) = \left[\prod_{i=1}^{I} \lambda_{d_i}^*(t_i)\right] \cdot \left[\prod_{i=0}^{I} \exp\left(-\int_{t_i}^{t_{i+1}} \lambda^*(u) \, du\right)\right].
$$

Simplifying the product of exponentials, we obtain:

$$
\mathcal{L}(\bm{s}) = \left[\prod_{i=1}^{I} \lambda_{d_i}^*(t_i)\right] \exp\left(-\int_{0}^{T} \lambda^*(u) \, du\right).
$$

This derivation demonstrates the relationship between the conditional intensity function and the likelihood function of a temporal point process.
\subsection{Derivation of the Posterior of TP$^2$DP$^2$}

Next, we derive the joint probability formula for our designed mixture model. Let us denote a mixture model where each event sequence $\bm{s}$ follows a mixture density
$
\sum_{m=1}^{M} \pi_m \mathcal{L}(\bm{s} \mid \bm{\theta}_m).
$

Given $N$ event sequences $\bm{S} = \{\bm{s}_n\}_{n=1}^{N}$, we denote the cluster allocation variables of each sequence as $\bm{c} = [c_1, \ldots, c_N] \in \{1, \ldots, M\}^N$. The joint distribution of all variables can be derived as follows:

\begin{eqnarray*}
\begin{aligned}
&p(M, \bm{\Theta}, \bm{\pi}, \bm{c}, \bm{S}) = p(M) p(\bm{\Theta}  \mid  M) p(\bm{\pi}  \mid  M) \prod_{n=1}^{N} \left[ \pi_{c_n} \mathcal{L}(\bm{s}_n  \mid  \bm{\theta}_{c_n}) \right] \\
&\approx p(M) p(\bm{\Theta}_S  \mid  M) p(\bm{\pi}  \mid  M)  \prod_{n=1}^{N} \left[ \pi_{c_n} \mathcal{L}(\bm{s}_n  \mid  \bm{\theta}_{S, c_n}, \widehat{\bm{\theta}}_{R, c_n}) \right] \times \delta(\bm{\Theta}_R - \widehat{\bm{\Theta}}_R) \\
&= p(M) p(\bm{U}  \mid  M) p(\bm{W}  \mid  M) p(\bm{\pi}  \mid  M)\prod_{n=1}^{N} \pi_{c_n} \mathcal{L}(\bm{s}_n  \mid  \bm{\theta}_{S, c_n}, \widehat{\bm{\theta}}_{R, c_n}) \\
&= p(M) p(\bm{U}  \mid  M) p(\bm{W}  \mid  M) p(\bm{\pi}  \mid  M)\prod_{n=1}^{N} \pi_{c_n} \left( \prod_{i=1}^{I_n} \lambda_{d_i,n}^*(t_i,n) \exp\left(-\int_{0}^{T_n} \lambda^*(u) \, du\right) \right),
\end{aligned}
\end{eqnarray*}

where $M$ is the number of clusters, $\bm{\Theta} = \{\bm{\theta}_m\}_{m=1}^{M} \in \mathbb{R}^P$ represents the collection of all parameters in component TPPs, $\bm{\pi} = [\pi_1, \ldots, \pi_M]$ specifies the probability of each cluster component, and $\mathcal{L}(\bm{s} \mid \bm{\theta}_m)$ is the likelihood of the event sequence given the parameters $\bm{\theta}_m$ of the $m$-th TPP. The terms $p(M)$, $p(\bm{\Theta}  \mid  M)$, and $p(\bm{\pi}  \mid  M)$ are the prior distributions of $M$, $\bm{\Theta}$, and $\bm{\pi}$, respectively. The conditional intensity function $\lambda_{d_i,n}^*(t_i,n)$ is for event type $d_i$ at time $t_i$ in the $n$-th sequence. The terms $\bm{\theta}_{S, c_n}$ and $\widehat{\bm{\theta}}_{R, c_n}$ are the subnetwork and remaining parameters, respectively.

Therefore, we have the joint posterior density function of TP$^2$DP$^2$:
\begin{eqnarray*}
\begin{aligned}
& p(M, \bm\Theta, \bm\pi, \bm{c}  \mid  \bm{S} ) \propto p(M)p(\bm\Theta \mid M) p(\bm{\pi}  \mid  M)  \sideset{}{_{n=1}^N}\prod \pi_{c_n} \mathcal{L}(\bm{s}_n  \mid  \bm{\theta}_{S,c_n}, \widehat{\bm\theta}_{R,c_n}) \\
&= p(M)p(\bm{U}  \mid  M)p(\bm{W} \mid M) p(\bm{\pi}  \mid  M) \sideset{}{_{n=1}^N}\prod \pi_{c_n} \left( \prod_{i=1}^{I_n} \lambda_{d_i,n}^*(t_i,n) \exp\left(-\int_{0}^{T_n} \lambda^*(u) \, du\right) \right).
\end{aligned}
\end{eqnarray*}

\section{Measure Theoretical Details and Construction of DPP Priors}\label{suppb}
We first elaborate on the measure theoretical details related to the DPP prior $\bm \mu$. Let $\Omega=\cup_{m=0}^{\infty} \Omega_m$ be the space of all finite subsets of $\mathbb R^q$ where
$
\Omega_m=\left\{\left\{\bm \mu_1, \ldots, \bm \mu_m\right\} \right\}
$
denotes the space of all finite subsets of cardinality $m$, $q$ is the dimension of each $\bm\mu_h$, and 
$\Omega_0=\{\emptyset\}$. In order to define a measure on the set of sets, we need to induce a well-defined measure by taking advantage of the measure on the set of real numbers. It is feasible to construct mappings, which map pairwise distinct $\left(\bm{\mu}_1, \ldots, \bm{\mu}_m\right) \in \mathbb{R}^{q m}$ into $\left\{\bm{\mu}_1, \ldots, \bm{\mu}_m\right\} \in \Omega_m$, and equip each $\Omega_m$ with the smallest $\sigma$-algebra making the mapping measurable. The $\sigma$-algebra $\mathcal{F}$ defined in the sample space $\Omega$ is the smallest $\sigma$-algebra that includes the union of $\sigma$-algebras defined on each set $\Omega_m$. Then, the measure on $\Omega$ is defined as follows: For sets $B = \cup_{m=0}^{\infty} B_m$ where $B_m \subseteq \Omega_m$,
\begin{equation}
    \int_B \mathrm{~d} \bm{U}=\sum_{m=0}^{\infty} \frac{1}{m !} \int_{B_m} \mathrm{~d} \boldsymbol{\mu}_m.
    \label{ap1}
\end{equation}
The presence of factorial in Eq.\ref{ap1} arises from the unordered nature of sets. 
For the construction of the DPP prior, we follow the method in~\cite{lavancier2015determinantal}. For mixture of Hawkes processes, we assume that the domain of the kernel function is the compact region related to the average intensity $\Bar{\lambda}$, with each dimension being the rectangle ranging from $[\Bar{\lambda}/2,  \Bar{\lambda} \times 2]$, where $\Bar{\lambda}$ is calculated by dividing the total number of events by time for mixture of Hawkes processes in the experiments. For neural TPPs, this range is set to a rectangle that can sufficiently encompass the point estimates obtained after the model pretraining stage. The way of spectral decomposition and approximation is similar to~\cite{bianchini2020determinantal} and the number of basis functions for the eigenfunctions approximation is set to 5.

\section{Conditional Gibbs Sampler for TP$^2$DP$^2$}\label{suppc}
we provide the comprehensive summary of the complete TP$^2$DP$^2$ posterior sampling procedure in Algorithm~\ref{alg:algorithm}.
\begin{algorithm}[h!t]
    \caption{Conditional Gibbs Sampler for TP$^2$DP$^2$}
    \label{alg:algorithm}
    \textbf{Input}: Event sequences $\bm{S}$, priors, initialization of the cluster number, maximum number of iteration $\mathbf{T}$, number of burn-in, step sizes for each update, point estimates $\widehat{\bm{\Theta}}_R$. \\
    \textbf{Output}: Posterior samples of variables in the model $\{M, \bm U, \bm r, \bm W, \bm c\}$.
    \begin{algorithmic}[1]
        \STATE Initialize parameters and set $j=0$.
        \WHILE{convergence not reached and $j < \mathbf{T}$}
            \STATE Sample non-allocated variables $(\bm{U}^{(na)}, \bm r^{(na)}, \bm W^{(na)})$ using collapsed Gibbs sampler. The sampling for $\bm{U}^{(na)}$ is given by:
            \begin{eqnarray*}
             \begin{aligned}
                 & p(\bm U^{(na)}  \mid \bm U^{(a)}, \bm r^{(a)}, \bm W^{(a)}, \bm c, u, \bm{S}) 
                 = p(\bm{U}^{(a)}\cup\bm{U}^{(na)}) \psi(u)^l,
            \end{aligned}
            \end{eqnarray*}
            and the sampling for $\bm{r}^{(na)}$ and $\bm{W}^{(na)}$ is given by:
            \begin{equation*}
            \begin{aligned}
               &p(\bm r^{(na)}  \mid  \cdots) \propto \sideset{}{_{m=1}^l}\prod p(r^{(na)}_m) e^{-u r^{(na)}_m},  \\ 
               &p(\bm W^{(na)}  \mid  \cdots) \propto \sideset{}{_{m=1}^l}\prod p(\bm w^{(na)}_m).
            \end{aligned}
            \end{equation*}
           \STATE Sample allocated variables $(\bm{U}^{(a)}, \bm r^{(a)}, \bm W^{(a)})$.
            \begin{eqnarray*}
            \begin{aligned}
                     p(\bm{U}^{(a)}  \mid  \cdots) &\propto p(\bm U^{(a)} \cup \bm U^{(na)} )  \sideset{}{_{m=1}^k}\prod \sideset{}{_{i: c_i = m}}\prod \mathcal{L}(\bm s_i  \mid  (\bm{\mu}^{(a)}_m, \bm w^{(a)}_m, \widehat{\bm\theta}_{R,m}),\\
                p(\bm r^{(a)} \mid \cdots) &\propto \sideset{}{_{m=1}^k}\prod p(r^{(a)}_m) (r^{(a)}_m)^{n_m}  \exp(-u r^{(a)}_m). \\
                p(\bm W^{(a)} \mid \cdots) &\propto \sideset{}{_{m=1}^k}\prod p(\bm w^{(a)}_m)  \sideset{}{_{i: c_i = m}}\prod \mathcal{L}(\bm s_i \mid (\bm{\mu}^{(a)}_m,\bm w^{(a)}_m, \widehat{\bm\theta}_{R,m})).
            \end{aligned}
            \end{eqnarray*}
            \STATE Sample cluster labels $\bm c$ using full conditional distribution:
            \begin{eqnarray*}
            \begin{aligned}
                p(c_i = m \mid \cdots) \propto
                \begin{cases}
                   & r^{(a)}_m \mathcal{L}(\bm s_i  \mid  \bm{\mu}^{(a)}_m,\bm w^{(a)}_m, \widehat{\bm\theta}_{R,m})  \quad \text{for~}m=1,...,k,\\
                   &  r^{(na)}_m \mathcal{L}(\bm s_i  \mid  \bm{\mu}^{(na)}_m,\bm w^{(na)}_m, \widehat{\bm\theta}_{R,m})  \quad \text{for~}m=k+1,...,k+l.
                \end{cases}
            \end{aligned}
            \end{eqnarray*}
            \STATE Update ancillary variable $u$ by $u \sim \text{Gamma}(N, \frac{1}{t})$.
            \STATE Increment $j$.
        \ENDWHILE
    \end{algorithmic}
\end{algorithm}

\section{Additional Experiment Details}\label{suppd}

\subsection{Additional Details of Experiments on Synthetic Mixture of Hawkes Processes Datasets}

We use the same parameterization of the components for both DMHP and TP$^2$DP$^2$, as in~\cite{NIPS2017_dd8eb9f2}. For the Hawkes process model which constitutes the $m$-th cluster component, its intensity function of the type-$d$ event at time $t$ is given by

\begin{equation}
\begin{aligned}
        \lambda_d^m(t) &=\bm{\mu}_d^m+\sum_{t_i<t} f_{d d_i}^m\left(t-t_i\right) \\ &=\bm{\mu}_d^m+\sum_{t_i<t} \sum_{j=1}^{N_m} a_{d d_i j}^m g_j\left(t-t_i\right),
\end{aligned}
\end{equation}
where $N_m$ is the number of basis functions for triggering functions, $a_{d d_i j}^m$ is the infectivity coefficient of $j$-th basis function $g_j$, deciding the infectivity of previous type-$d_i$ event on the current event. We set all basis functions to be the Gaussian kernel functions and set them to be the same for both DMHP and TP$^2$DP$^2$.

\subsection{Additional Details of Experiments on Synthetic Mixture of Hybrid Point Processes Datasets and Real-World Datasets}

\subsubsection{Synthetic Mixture of Hybrid Point Processes Dataset Construction} We use the following point processes to generate event sequences, and mix the obtained event sequences to construct a dataset: 1) Homogeneous Poisson process (Homo). This process involves events that occur at a constant rate independently of each other. 2) Inhomogeneous Poisson process (Inhomo). In this process, the event rate varies over time, allowing for a non-constant rate of occurrence. 3) Self-correcting process (Sc). Within this process, the occurrence of one event inhibits the others according to the self-correcting mechanism. 4) Hawkes process (Hks). The occurrence of one event triggers the likelihood of subsequent events, which manifests the self-exciting characteristic. 5) neural point process (Npp): we utilize the Neural Hawkes Process (NHP)~\cite{mei2017neural}, a neural network model that can flexibly capture complex sequential dependencies between events (excitatory, inhibitory, time-varying, etc.) to generate sequences. By randomizing the parameters of each network layer, we obtain an instantiation of NHP. Subsequently, the Ogata’s modified thinning algorithm ~\cite{ogata1981lewis} is applied to simulate the event sequence based on the intensity function modeled by the NHP.
 
The mixing scheme of various types of point processes in different datasets is as follows: Homo + Inhomo + Hks for ${K_{GT}} = 3$, Homo + Inhomo + Hks  + Sc for ${K_{GT}} = 4$, and Homo + Inhomo + Hks  + Sc + Npp for 
${K_{GT}} = 5$.

\subsubsection{Training details of Experiments on Synthetic Mixture of Hybrid Point Processes Datasets and Real-World Datasets}

In a mixture model of point processes, several hyperparameters need to be specified with special care. These hyperparameters include: 1) hyperparameters in the backbone model structure, such as the number of layers, hidden sizes, etc. 2) hyperparameters of the inference algorithm, such as the number of SGD updates in the M-step of the Dirichlet mixture framework, the variance of the proposal distribution in the TP$^2$DP$^2$ model, etc. 3) other regular parameters like the batch size and learning rate.  We tune each model in the both frameworks to the best state according to performance in the validation set, in which grid search is used. For the posterior inference of TP$^2$DP$^2$, the initialization of parameters is naturally obtained via the maximum likelihood estimation achieved by stochastic gradient descent during the pretraining stage. The Bayesian subnetwork incorporates several layers as a whole of the full model, rather than being selected in the parameter level. Whether a certain layer is inferred via the Bayesian approach or kept fixed at point estimates is determined empirically by traversing the possible combinations, and the subnetwork that performs the best in the validation set is adopted. In the experiments, we find that making the last 1-2 fully connected layers (which serve as an aggregator of historical and current information) in neural point processes to be the Bayesian subnetwork generally attains satisfying and robust results in most scenarios. For the step size in each update, it is useful to make it decay as the algorithm iterates.

\section{Additional Exmerimental Results}\label{suppe}

\subsection{Additional Exmerimental Results on Synthetic Mixture of Hawkes Processes Datasets}
In Figure~\ref{fig:tsne_syn1_true}, each panel presents a t-SNE plot of the Hawkes processes with different base intensities, and different panels illustrates different levels of base intensity disparities across clusters. The ground truth label is illustrated with different colors. It can be seen that when the disparity in the true base intensity among different point processes is minimal (as seen in the upper two subfigures), the inherent distinctions within event sequences are not apparent. In such cases, clustering them into excessive categories is futile. 
\begin{figure}[t]
    \centering
    \includegraphics[width=0.6\linewidth]{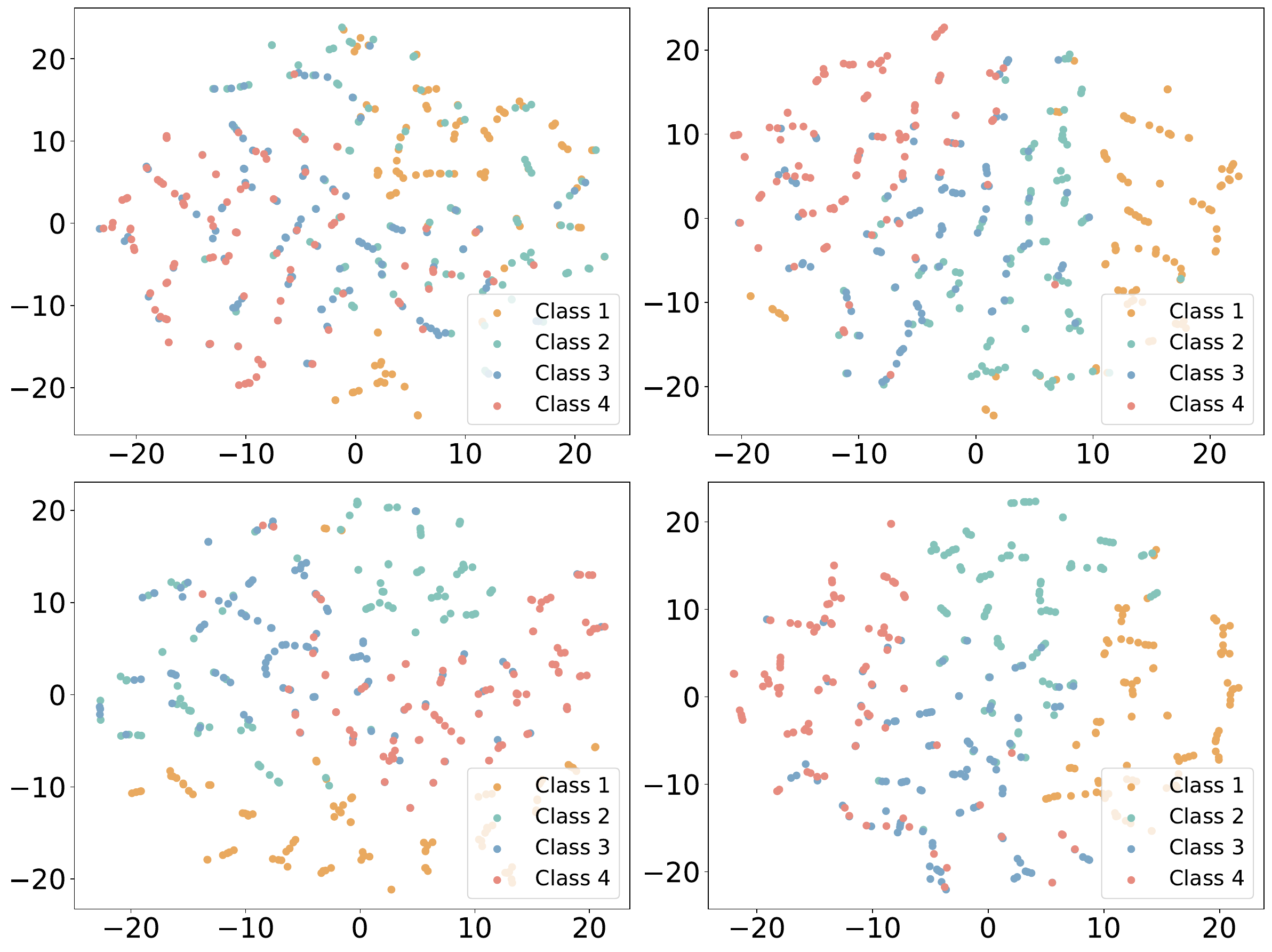}
    \caption{The t-SNE plot of the ground truth distribution for  synthetic mixture of Hawkes processes datasets with $\delta$ values of 0.2 (upper left), 0.4 (upper right), 0.6 (lower left), and 0.8 (lower right).}
    \label{fig:tsne_syn1_true}
\end{figure}

\subsection{Additional Exmerimental Results on Synthetic Mixture of Hybrid Point Processes Datasets}

\begin{figure}[h]
    \centering
\includegraphics[width=0.6\linewidth]{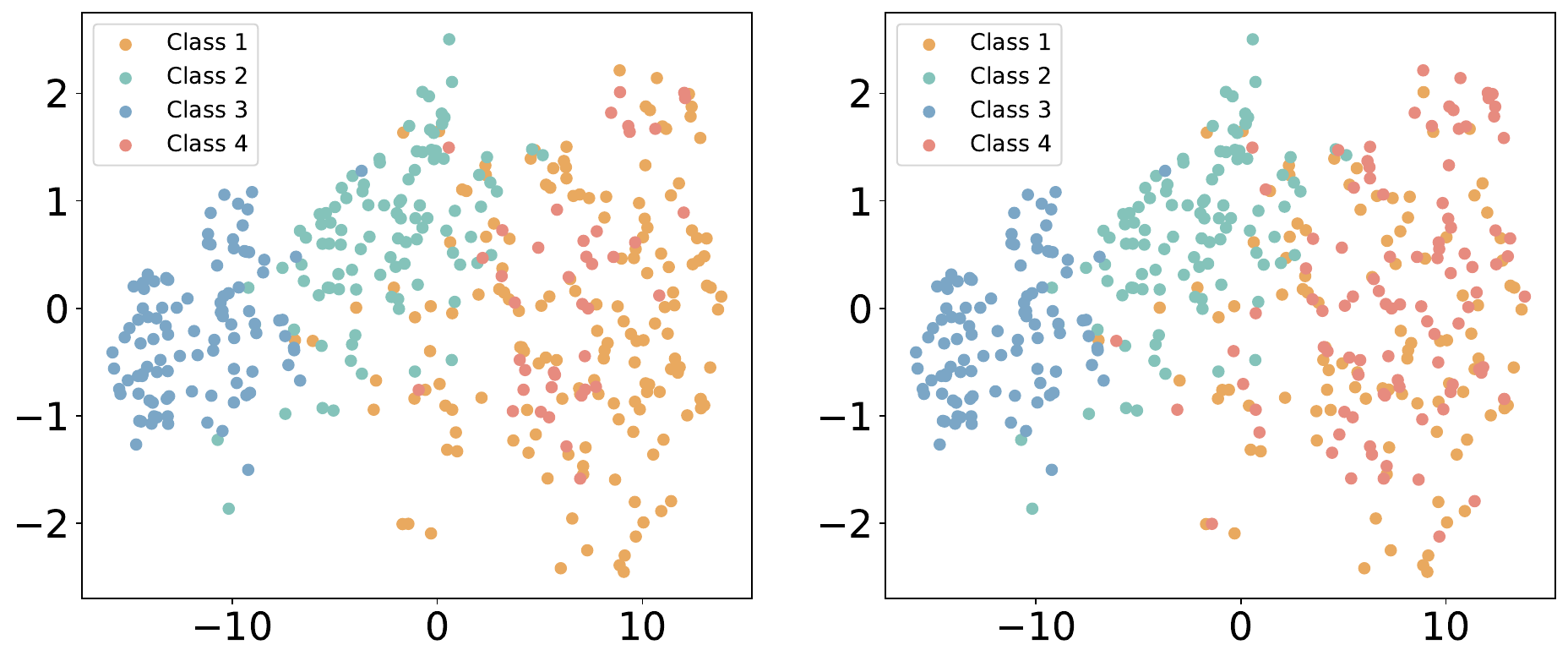}
    \caption{The t-SNE plot of the clustering result yielded by NTPP-MIX (left) and TP$^2$DP$^2$ (right) with RMTPP backbone for the synthetic mixture of hybrid point processes datasets which has four clusters.}
    \label{fig:tsne_syn2_k4}
\end{figure}

\begin{figure}[!htb]
    \centering
    \includegraphics[width=0.6\linewidth]{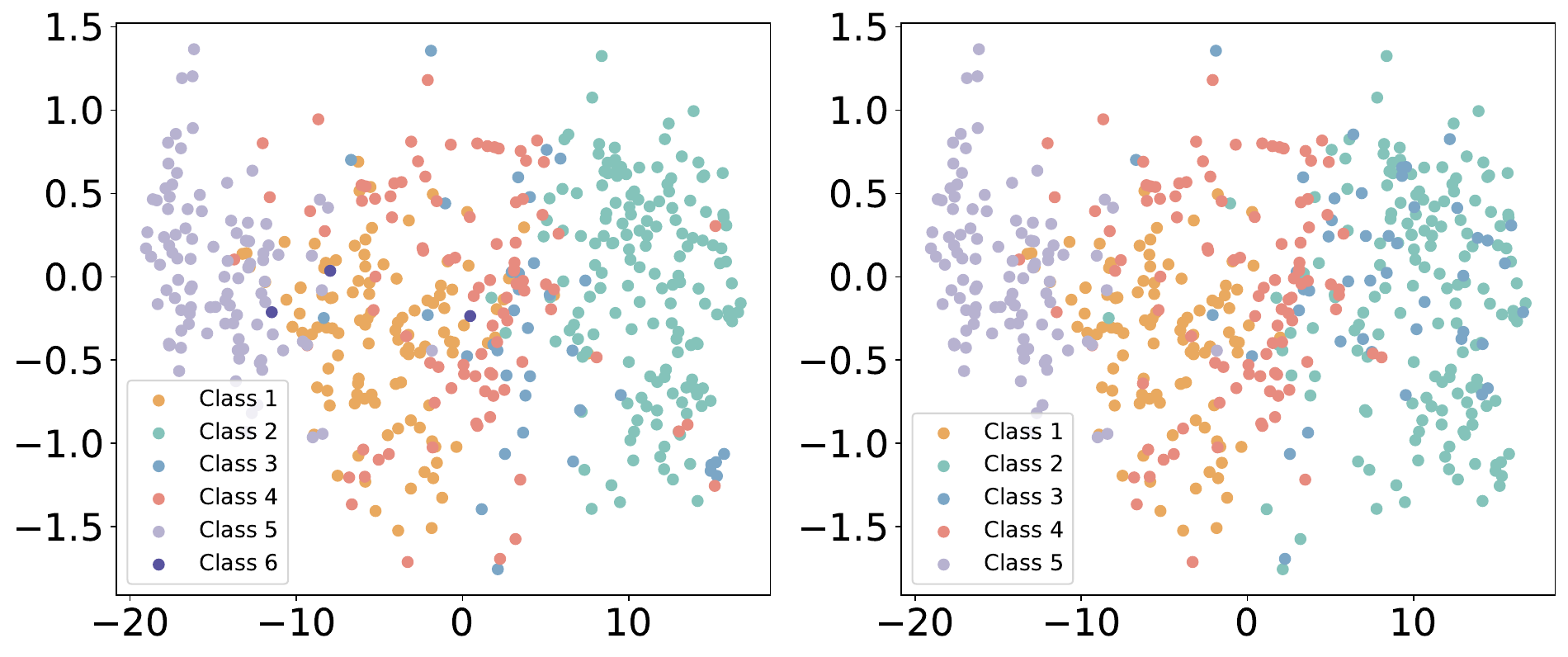}
    \caption{The t-SNE plot of the clustering result yielded by NTPP-MIX (left) and TP$^2$DP$^2$ (right) for the synthetic mixture of hybrid point processes datasets which has five clusters.}
    \label{fig:tsne_syn2_k5}
\end{figure}

Figure~\ref{fig:tsne_syn2_k4} and Figure~\ref{fig:tsne_syn2_k5} show t-SNE plots of the clustering results for mixture of four and five hybrid point processes, respectively. TP$^2$DP$^2$ achieves higher mean clustering purity scores in both cases.
\clearpage

\end{document}